\documentclass[lettersize,journal]{IEEEtran}

\usepackage{graphicx}
\usepackage{array}
\usepackage{makecell}
\usepackage{multirow}
\usepackage{amsmath,amsfonts}
\usepackage[normalem]{ulem}
\usepackage{tcolorbox}
\usepackage{colortbl}
\usepackage{enumitem}
\usepackage{url}
\usepackage{enumerate}
\usepackage{booktabs}
\usepackage{xspace}

\usepackage{color}
\usepackage{threeparttable}
\usepackage{ulem}
\usepackage{amssymb}

\usepackage{float}
\usepackage[ruled,linesnumbered]{algorithm2e}
\usepackage{algorithm2e, setspace, algpseudocode}
\usepackage{balance}
\usepackage[caption=false,font=normalsize,labelfont=sf,textfont=sf]{subfig}
\usepackage{textcomp}
\usepackage{stfloats}
\usepackage{verbatim}
\usepackage{balance}
\usepackage{hyperref}

\def\BibTeX{{\rm B\kern-.05em{\sc i\kern-.025em b}\kern-.08em
    T\kern-.1667em\lower.7ex\hbox{E}\kern-.125emX}}

\definecolor{green_bg}{RGB}{180, 215, 183}
\definecolor{green_hyp}{RGB}{160, 196, 143}

\newcommand{\eg}{\textit{e.g.,}\xspace}
\newcommand{\ie}{\textit{i.e.,}\xspace}

\newtheorem{definition}{Definition}
\newcommand{\BasicTS}{BasicTS+\xspace}

\newtcolorbox{mybox}[3][]
{
  colframe = #2!100,
  colback  = #2!30,
  #1,
}

\begin{document}

\title{Exploring Progress in Multivariate Time Series Forecasting: Comprehensive Benchmarking and Heterogeneity Analysis}

\author{
  Zezhi Shao, Fei Wang, Yongjun Xu, Wei Wei, Chengqing Yu, Zhao Zhang, Di Yao, Tao Sun, \\Guangyin Jin, Xin Cao, Gao Cong, Christian S.Jensen, Xueqi Cheng
\thanks{
Zezhi Shao, Fei Wang, Yongjun Xu, Chengqing Yu, Zhao Zhang, Di Yao, Tao Sun, Xueqi Cheng are with the Institute of Computing Technology, CAS, Beijing 100190, China.
Zezhi Shao and Chengqing Yu are also with the University of Chinese Academy of Sciences, Beijing 100049, China.
(e-mail: \{shaozezhi19b, wangfei, xyj, yuchengqing22b, zhangzhao2021, yaodi, suntao, cxq\}@ict.ac.cn)
}
\thanks{
Wei Wei is with the School of Computer Science and Technology, Huazhong University of Science and Technology, Hubei 430074, China
(e-mail: weiw@hust.edu.cn).
}
\thanks{
Guangyin Jin is with the Tsinghua University, Beijing, China.
(e-mail: jinguangyin96@foxmail.com)}
\thanks{
Xin Cao is with the School of Computer Science and Engineering, The University of New South Wales, Australia.
(e-mail: xin.cao@unsw.edu.au).
}
\thanks{
Gao Cong is with the School of Computer Science and Engineering, Nanyang Technological University, Singapore.
(e-mail: gaocong@ntu.edu.sg).
}
\thanks{
Christian S.Jensen is with the Department of Computer Science, Aalborg University, Denmark.
(e-mail: csj@cs.aau.dk).
}
\thanks{Corresponding Authours: Fei Wang (wangfei@ict.ac.cn), Yongjun Xu (xyj@ict.ac.cn), and Xueqi Cheng (cxq@ict.ac.cn).}
}

\markboth{IEEE TRANSACTIONS ON KNOWLEDGE AND DATA ENGINEERING}
{}

\maketitle

\begin{abstract}
Multivariate Time Series (MTS) analysis is crucial to understanding and managing complex systems, such as traffic and energy systems, and a variety of approaches to MTS forecasting have been proposed recently.
However, we often observe inconsistent or seemingly contradictory performance findings across different studies.
This hinders our understanding of the merits of different approaches and slows down progress.
We address the need for means of assessing MTS forecasting proposals reliably and fairly, in turn enabling better exploitation of MTS as seen in different applications.
Specifically, we first propose \BasicTS, a benchmark designed to enable fair, comprehensive, and reproducible comparison of MTS forecasting solutions.
\BasicTS establishes a unified training pipeline and reasonable settings, enabling an unbiased evaluation.
Second, we identify the heterogeneity across different MTS as an important consideration and enable classification of MTS based on their temporal and spatial characteristics. Disregarding this heterogeneity is a prime reason for difficulties in selecting the most promising technical directions.
Third, we apply \BasicTS along with rich datasets to assess the capabilities of more than 45 MTS forecasting solutions. This provides readers with an overall picture of the cutting-edge research on MTS forecasting.
The code can be accessed at {\color{blue}\url{https://github.com/GestaltCogTeam/BasicTS}.}

\end{abstract}


\section{Introduction}

\IEEEPARstart{S}{ensors} are increasingly being deployed in complex, real-world systems. Readings from such sensors form Multivariate Time Series (MTS) that in turn are used for understanding and operating the host systems.
For instance, the PEMS~\cite{PEMS-BAY} dataset consists of traffic data from critical locations in a transportation system, and the Electricity~\cite{2018LSTNet} dataset records the electricity consumption by key clients in a power system. Consequently, MTS forecasting has become fundamental to understanding and operating complex real-world systems, enabling applications such as traffic management~\cite{2022D2STGNN}, emergency management~\cite{Nature}, and resource optimization~\cite{2023SUFS}. 

MTS data analysis must consider both the temporal and spatial aspects of the data~\cite{2018DCRNN, 2021Informer}.
The temporal aspect often encompasses complex dynamics, including non-stationarity, periodicity, and randomness. 
The spatial aspect concerns interdependencies among time series, known as spatial dependencies~\cite{2018DCRNN} or cross-dimension dependencies~\cite{2023Crossformer}, which can affect prediction accuracy substantially.
Effective modeling the complex temporal and spatial aspects of MTS is a key challenge, which also has been addressed in many studies.

Recent MTS forecasting solutions have been based predominantly on deep learning~\cite{2021Informer, 2021AutoFormer, 2022FEDFormer, 2018DCRNN, 2019GWNet, 2018STGCN}.
These solutions often address two prominent and more specific problems, namely {Long-term Time Series Forecasting~(LTSF)} and {Spatial-Temporal Forecasting~(STF)}, in which the modeling of temporal and spatial patterns in the data are essential.
{\color{black}
{LTSF} solutions are concerned with long-term forecasting and often employ advanced neural networks like Transformers~\cite{2017Transformer} to model long-term temporal dependencies.
Notable solutions include efficient Transformers~\cite{2019LogTrans, 2021Informer, 2022Pyraformer}, series-level correlations~\cite{2021AutoFormer}, frequency-based solutions~\cite{2022FEDFormer}, and Transformers utilizing patched time series~\cite{2023PatchTST, 2023Crossformer}.}
In contrast, STF solutions aim to improve prediction by effectively modeling spatial correlations. The prevalent approach is to combine Graph Convolution Networks~(GCN)~\cite{2017GCN} with different sequence models~\cite{2014GRU, 2016TCN} to form Spatial-Temporal Graph Neural Networks~(STGNN).
Examples include combining GCNs with Recurrent Neural Networks~(RNN)~\cite{2018DCRNN}, Convolutional Neural Networks~(CNN)~\cite{2019GWNet}, and Attention mechanism~\cite{2020GMAN, HUTFormer}

While proposals of new solutions include experimental studies, such studies are at times incomparable or seemingly inconsistent. This causes uncertainty on which directions to take and impedes progress towards better solutions. 
As an example of the current state of affairs, some studies~\cite{2020STSGCN, 2020StemGNN, 2021Z-GCNets, STGODE} report poor performance of the key baselines DCRNN~\cite{2018DCRNN} and GWNet~\cite{2019GWNet}, at up to 33\% lower than the performance we reproduce. 
Next, proposals of LTSF solutions~\cite{2021Informer, 2021AutoFormer, 2022FEDFormer, 2023Crossformer} usually report evaluations solely using metrics like MAE and MSE based on normalized time series, making prediction errors seem to be very low. An alternative is to perform evaluations on re-normalized data and to report more metrics like MAPE and WAPE, which are not affected by the range of data.
Issues such as these prevent researchers from judging the strengths and weaknesses of different solutions.

{\color{black}Further, some studies present seemingly contradictory findings in selecting which technical directions to take when pursuing better solutions to LTSF and STF.}
In relation to the temporal aspect, {\color{black}(i)~\textit{\underline{the effectiveness of advanced} \underline{neural networks has been debated}}~\cite{2021Informer, 2023DLinear, 2023PatchTST, DSformer}.
One study~\cite{2023DLinear} finds that LTSF-Linear, which employs a simple linear layer, significantly outperforms Transformer-based models~\cite{2021Informer,2021AutoFormer,2022FEDFormer,2022Pyraformer}, and the study concludes that Transformer-based architectures are not as effective as previously claimed.
However, subsequent studies~\cite{2023PatchTST, 2022TimesNet,DSformer} find that advanced neural networks outperform LTSF-Linear.
We find that the difference in model size between these approaches makes it difficult to determine their relative effectiveness.}
In relation to the spatial aspect, (ii)~\textit{\underline{the necessity of GCNs has been questioned}}~\cite{2022SimST, 2022STID}.
While STGNNs have brought significant improvements, many recent studies highlight the inefficiency of STGNNs and explore alternative means of modeling the dependencies among time series, \eg normalization~\cite{2021STNorm}~\cite{2022STID}.
The success of these non-GCN methods indicates the need for a deeper understanding of spatial dependencies and for insight into when these alternative methods are effective.

To mitigate issues such as those exemplified above and to offer insight into the advance achieved, we contribute a comprehensive analysis and comparison of both MTS forecasting datasets and models.
First, as we believe that providing a fair, comprehensive, and reproducible benchmark for MTS forecasting can mitigate the current state of affairs and enable progress, we introduce \BasicTS, a benchmark for studying and comparing MTS forecasting solutions. \BasicTS establishes a unified training pipeline and reasonable evaluation settings. The former resolves inconsistent performance issues caused by unique data and experimental setups in previous studies while the latter enables a more intuitive evaluation of prediction errors.
Overall, \BasicTS facilitates a fair, comprehensive, and reproducible evaluation of over 45 popular MTS forecasting solutions on 20 commonly used datasets.{\color{black}\footnote{Due to space limitations, not all baselines and datasets are presented in this paper.}}

Second, we address the problem of selecting an appropriate technical approach by studying the impact of the heterogeneity across MTS datasets. We use heterogeneity to refer to completely different patterns observed across different MTS datasets. 
In the temporal aspect, we classify datasets into those with stable patterns, significant distribution drift, and unclear patterns. 
In the spatial aspect, we find that spatial sample indistinguishability is a key concept and partition datasets into those with and without significant spatial sample indistinguishability. 
Experimental studies show that previous conclusions are valid only for certain types of data. 
{\color{black}For example, basic neural networks~\cite{2023DLinear} only outperform advanced neural networks~\cite{2021Informer,2021AutoFormer,2022FEDFormer} on datasets without stable temporal patterns,} and approaches for modeling spatial dependencies, such as GCN-based approaches, are only effective on datasets with significant spatial sample indistinguishability. 
We find that blindly adopting conclusions from previous studies can lead researchers to make misguided inferences.

{\color{black}
Moreover, by using \BasicTS with heterogeneous datasets, we conduct an exhaustive analysis and comparison of popular solutions. Initially, we discuss how to design or select MTS prediction solutions for a given MTS dataset, as well as how to choose suitable datasets for evaluating a given MTS forecasting solution. Subsequently, we present detailed experimental results on the performance and efficiency of popular solutions across comprehensive datasets, shedding light on the advancements made.
The objective of these results and discussions is to accelerate progress and facilitate researchers in drawing more reliable conclusions.
Additionally, we highlight directions that deserve more attention. }
In summary, we make the following main contributions:

\begin{itemize}[leftmargin=25pt]
    \item 
    We present \BasicTS, the first benchmark specifically designed for fair comparison of MTS forecasting  solutions, especially both STF and LTSF solutions. 
    {\color{black}\BasicTS facilitates evaluation of over 45 popular models on 20 datasets to address the seemly inconsistent performance findings.}

    \item 
    We identify heterogeneity among MTS datasets as a key challenge, and classify datasets based on temporal and spatial characteristics.
    We find that neglecting heterogeneity is a cause of {\color{black}difficulties in selecting technical directions}, and that previous conclusions apply only to certain types of data. 

    \item 
    We conduct an extensive analysis and comparison of popular models using \BasicTS together with rich heterogeneous datasets. 
    The findings offer valuable insight into the progress already made, aiding researchers in choosing appropriate solutions or datasets, and drawing more reliable conclusions.
\end{itemize}

The paper is organized as follows.
Section~\ref{sec:related_work} provides discussions of related work on LTSF, STF, and MTS forecasting benchmarking.
Section~\ref{sec:preliminaries} covers preliminaries and essential definitions.
Section~\ref{section:benchmark_building} presents the \BasicTS benchmark.
Section~\ref{sec:het_data} then delves into the heterogeneity among MTS datasets, and provides hypotheses for explaining seemingly {\color{black}contradictory} findings.
Section \ref{sec:experiments} reports on the application of \BasicTS to popular models and provides new insights.
Section \ref{sec:conclusions} concludes the paper.

\section{Related Work}
\label{sec:related_work}
We cover studies related to LTSF and STF, which are the two most prominent topics in recent MTS forecasting studies. We present their goals, techniques, and related open issues. Furthermore, we cover existing MTS benchmarking studies.

\subsection{Long-term Time Series Forecasting}

To achieve accurate long-term time series forecasting~\cite{LTSF_Survey}, studies concentrate on capturing the temporal patterns in MTS data, and have proposed methods to efficiently and effectively incorporate longer-term historical information. 
{\color{black}For example, forecasting future electricity demand over several months or even years in power systems is a typical application scenario, where such predictions are crucial for resource optimization and strategic planning.}

Early studies typically propose traditional statistical methods~(\eg ARIMA~\cite{ARIMA} and ETS~\cite{ETS}) or machine learning methods~(\eg GBRT~\cite{GBRT} and SVR~\cite{SVR}).
These methods often struggle to handle high non-linearity well, and they typically rely heavily on stationarity-related assumptions~\cite{2022D2STGNN}.
{\color{black}With the advent of deep learning~\cite{innovation, innovation2}, studies have embraced more powerful and advanced neural architectures for time series modeling, such as TCN~\cite{2016TCN}, LSTM~\cite{2014Seq2Seq}, and Transformer~\cite{2017Transformer}.}
Among these, Transformer-based models have garnered increasing attention.
Informer~\cite{2021Informer} proposes a ProbSparse self-attention mechanism and distilling operation to address the quadratic complexity of the Transformer, leading to significant performance improvements and being recognized as a milestone in LTSF~(AAAI 2021 best paper).
Subsequently, Autoformer~\cite{2021AutoFormer} features an efficient auto-correlation mechanism to discover and aggregate information at the series level, while FEDformer~\cite{2022FEDFormer} proposes an attention mechanism with low-rank approximation in frequency and a mixture of {\color{black}experts} to control distribution shifts.
Additionally, Pyraformer~\cite{2022Pyraformer} designs pyramidal attention to effectively describe short and long temporal dependencies with low complexity.
Overall, the Transformer architecture is widely regarded as one of the most effective and promising approaches for MTS forecasting.

However, a recent study proposes LTSF-Linear~\cite{2023DLinear} and questions
 the effectiveness of Transformer architectures. LTSF-Linear employs a simple linear layer and outperforms all the earlier models. 
It carefully examines every key components of Transformers and concludes that they are ineffective at time series forecasting. 
This conclusion has subsequently been challenged by studies~\cite{2023PatchTST, DSformer, 2022TimesNet} that employ {\color{black}advanced neural networks} to outperform LTSF-Linear.
Nevertheless, considering the substantial difference in model size and the small difference in predictive performance, understanding fully the effectiveness of {\color{black}advanced models} remains challenging. Furthermore, more exploration is required to understand why a simple linear model can achieve state-of-the-art performance.

\subsection{Spatial-Temporal Forecasting}
In contrast to LTSF, spatial-temporal forecasting must contend with not only temporal dynamics in time series but also dependencies among time series. 
{\color{black}
A prime example of this is in traffic management, where predicting future conditions requires data from multiple traffic sensors, clearly highlighting the spatial dependencies among these sensors. 
Consequently, considerable research has been devoted to effectively capture and model these spatial and temporal patterns.
}

Early deep learning approaches often employ CNNs to process spatial information and combine CNNs and RNNs~\cite{2015convLSTM, 2018DMVST, 2018LSTNet}.
However, as the relationships among time series are usually non-Euclidean, grid-based CNNs may not be optimal for handling spatial dependencies.
With the development of GCNs~\cite{2016GCN,2017GCN}, STGNNs~\cite{2018DCRNN, 2018STGCN} have gained increased attention.
STGNNs utilize GCNs to model spatial dependencies based on pre-defined prior graphs, and further combine them with sequential models~\cite{2016TCN,2014GRU,2017Transformer}.
For example, models like DCRNN~\cite{2018DCRNN}, ST-MetaNet~\cite{2019STMetaNet}, and DGCRN~\cite{2021DGCRN} incorporate GCNs with RNNs~\cite{2014GRU} and their variants, and then predict step by step following the seq2seq~\cite{2014Seq2Seq} architecture.
Graph WaveNet~\cite{2019GWNet}, STGCN~\cite{2018STGCN}, and Auto-DSTSGN~\cite{jin2022automated} integrate GCNs with gated TCNs and their variants to facilitate parallel computation.
Futhermore, attention mechanisms are used widely in STGNNs, such as GMAN~\cite{2020GMAN}, ASTGNN~\cite{2021ASTGNN}.
In addition, neural architecture search solutions~\cite{jin2022automated, nas1} have also received widespread attention.
However, many recent studies argue that the pre-defined prior graph might be biased, incorrect, or even unavailable in many cases.
Thus, they propose to jointly learn the graph structure~(\ie a latent graph) and optimize STGNNs, \eg AGCRN~\cite{2020AGCRN}, MTGNN~\cite{2020MTGNN}, StemGNN~\cite{2020StemGNN}, GTS~\cite{2021GTS}, DFDGCN~\cite{DFDGCN}, and STEP~\cite{2022STEP}.

However, both prior graph-based STGNNs and latent graph-based STGNNs are usually have a complexity ranging from $O(N^2L)$ to $O(N^2L^2)$ due to the graph convolution operation, where $N$ is the number of time series and $L$ is the length of a time series.
Consequently, recent studies~\cite{2023NonGCN1, 2023NonGCN2} have questioned the necessity of STGNNs~\cite{2021STNorm, 2022SimST, 2023GraphFree, STNorm2} and have explored alternative techniques~\cite{2022STID,2021STNorm,2022MSDR}. 
For instance, STNorm~\cite{2021STNorm} introduces spatial-temporal normalization, and STID~\cite{2022STID} implements a simple yet effective spatial-temporal identity attaching approach. 
These solutions achieve similar prediction performance as STGNNs but with significantly higher efficiency.  The success of these non-GCN solutions highlights the need for a deeper understanding of spatial dependencies and when and how these solutions are effective.

\subsection{MTS Forecasting Benchmarking}

Several benchmarking studies have been devoted to MTS forecasting and associated downstream tasks.
For example, studies like DGCRN~\cite{2021DGCRN}, LibCity~\cite{2021LibCity}, DL-Traff~\cite{2021DL-Traff}, and {\color{black}our previous work BasicTS~\cite{BasicTS}}, use the benchmarks to address STF-based downstream tasks, \eg urban spatial-temporal forecasting~\cite{jin2023spatio}.
Similarly, the studies that contribute LTSF-Linear~\cite{2023DLinear} and TimesNet~\cite{2022TimesNet} propose benchmarks for LTSF. 
However, these benchmarks have several limitations. 
First, they only cover some of the research in either STF or LTSF, and cannot address comprehensively the temporal and spatial aspects of MTS. 
Second, many of them lack a unified pipeline and instead train each baseline individually with a unique pipeline, {\color{black}which may lead to unfairness.}
Third, these benchmarks are incapable of covering adequately the issues related to the different technical approaches, to contending with the temporal and spatial aspects of MTS forecasting.

Notably, the motivation and contribution of this study significantly differ from \cite{BasicTS}. The focus of this study is to reliably and fairly evaluate MTS forecasting solutions, reveal the heterogeneity across MTS datasets, and address seemingly inconsistent findings in existing studies. This aims to enhance the utilization of MTS in various applications rather than solely proposing benchmarks, surpassing mere software-level contributions.
Moreover, even from a software perspective, \BasicTS has been refactored to adapt and apply to both STF and LTSF tasks (whereas BasicTS~\cite{BasicTS} is designed only for STF). \BasicTS also incorporates more extensible features.

\begin{table}[t]
\renewcommand\arraystretch{1}
\setlength\tabcolsep{1pt}
    \centering
    \caption{Inconsistent performance of GWNet and DCRNN in highly cited papers. The pink background marks the worst performance, while the green background marks the performance produced by \BasicTS. 
    Assuming that $x$ and $y$ are the values reported in previous studies and \BasicTS, respectively, then the gap is defined as $(x-y)/x \cdot 100\%$.}
    \scalebox{0.75}{
    \label{tab_inconsistent_performance}
    \begin{tabular}{cc|ccr|ccr}
      \toprule
      \midrule
      \multicolumn{1}{c}{\multirow{2}*{\textbf{}}} & \multicolumn{1}{c}{\multirow{2}*{\textbf{Source}}} & \multicolumn{3}{c}{\textbf{PEMS04}} & \multicolumn{3}{c}{\textbf{PEMS08}}\\ 
      \cmidrule(r){3-5} \cmidrule(r){6-8} 
       & & MAE & RMSE & MAPE & MAE & RMSE & MAPE \\
      \hline
      \midrule
      \multirow{5}*{\rotatebox{90}{\textbf{GWNet\quad\ }}}
      & {\small\cite{2020STSGCN, 2021STFGCN, 2020StemGNN, 2021SCINet, 2022MSDR}} & 25.45 & 39.70 & 17.29\% & 19.13 & 31.05 & 12.68\% \\
      & {\small\cite{STGODE, 2022STG-NCDE}} &24.89 & 39.66 & 17.29\% & 18.28 & 30.04 & 12.15\% \\
      & {\small\cite{2021ASTGNN}} &19.36 & 31.72 & 13.31\% & 15.07 & 23.85 & 9.51\% \\
      & {\small\cite{2021Z-GCNets}} &{\cellcolor{pink}28.15} & {\cellcolor{pink}39.88} & {\cellcolor{pink}18.52\%} & {\cellcolor{pink}20.30} & {\cellcolor{pink}30.82} & {\cellcolor{pink}13.84\%} \\
      & \BasicTS  &{\cellcolor{green_bg}\textbf{18.80}} & {\cellcolor{green_bg}\textbf{30.14}} & {\cellcolor{green_bg}\textbf{13.19\%}} & {\cellcolor{green_bg}\textbf{14.67}} & {\cellcolor{green_bg}\textbf{23.55}} & {\cellcolor{green_bg}\textbf{9.46\%}} \\
      \cmidrule(r){2-8}
      & Gap & {\color{teal}\textbf{33.21}\%$\uparrow$} & {\color{teal}\textbf{24.42}\%$\uparrow$} & {\color{teal}\textbf{28.78}\%$\uparrow$} &{\color{teal}\textbf{27.73}\%$\uparrow$} &{\color{teal}\textbf{23.59}\%$\uparrow$} &{\color{teal}\textbf{31.64}\%$\uparrow$} \\
      \hline
      \midrule
      \multirow{6}*{\rotatebox{90}{\textbf{DCRNN\quad}}}
      & {\small\cite{2020STSGCN, 2021STFGCN, 2020StemGNN, 2021SCINet, 2022MSDR}} &{\cellcolor{pink}24.70} & {\cellcolor{pink}38.12} & {\cellcolor{pink}17.12\%} & 17.86 & 27.83 & 11.45\% \\
      & {\small\cite{STGODE}} &24.63 & 37.65 & 17.01\% & 17.46 & 27.83 & 11.39\% \\
      & {\small\cite{2021ASTGNN}} &23.65 & 37.12 & 16.05\% & {\cellcolor{pink}18.22} & {\cellcolor{pink}28.29} & {\cellcolor{pink}11.56\%} \\
      & {\small\cite{2021Z-GCNets, 2020AGCRN, 2022STG-NCDE}} &21.22 & 33.44 & 14.17\% & 16.82 & 26.36 & 10.92\% \\
      & \BasicTS  & {\cellcolor{green_bg}\textbf{19.66}} & {\cellcolor{green_bg}\textbf{31.18}} & {\cellcolor{green_bg}\textbf{13.45\%}} & {\cellcolor{green_bg}\textbf{15.23}} & {\cellcolor{green_bg}\textbf{24.17}} & {\cellcolor{green_bg}\textbf{10.21\%}} \\
      \cmidrule(r){2-8}
      & Gap & {\color{teal}\textbf{20.40}\%$\uparrow$} & {\color{teal}\textbf{12.20}\%$\uparrow$} & {\color{teal}\textbf{21.43}\%$\uparrow$} & {\color{teal}\textbf{16.41}\%$\uparrow$} &{\color{teal}\textbf{14.56}\%$\uparrow$} &{\color{teal}\textbf{11.67}\%$\uparrow$} \\
      \midrule
      \bottomrule
    \end{tabular}
    }
\end{table}

\section{Preliminaries}
\label{sec:preliminaries}
We define key concepts and the forecasting task.

\begin{definition}
\textbf{Multivariate Time Series.}
A multivariate time series includes multiple time-dependent variables.
It can be expressed as a matrix $\mathbf{X}\in\mathbb{R}^{T\times N}$, where $T$ is the number of time steps and $N$ is the number of variables.
We additionally denote the data in time series $i$ ranging from $t_1$ to $t_2$ as $\mathbf{X}^i_{t_1:t_2}$.
\end{definition}

\begin{definition}
\textbf{Multivariate Time Series Forecasting.}
Given historical data $\mathbf{X}\in\mathbb{R}^{T_h\times N}$ from the past $T_h$ time steps, multivariate time series forecasting aims to predict $\mathbf{Y}\in\mathbb{R}^{T_f\times N}$ of the $T_f$ nearest future time steps.
\end{definition}
\section{Benchmark Construction}
\label{section:benchmark_building}

We present \BasicTS, a benchmark designed for fair, comprehensive, and reproducible evaluation of MTS forecasting solutions, including both STF and LTSF solutions.

\subsection{Unified Training Pipeline}
We proceed to delve into the root causes of seemingly inconsistent performance findings and propose in response a unified training pipeline, thereby enabling fair comparison of forecasting models.

\subsubsection{Inconsistent forecasting performance.} The inconsistencies imply that the forecasting performance of the same solution exhibits notable variations across experimental studies in different papers, even when on the same dataset and with the same experimental settings.
To illustrate this, Table \ref{tab_inconsistent_performance} compiles performance findings from studies in a range of papers for two solutions that are often used as baselines: DCRNN~\cite{2018DCRNN} and Graph WaveNet~\cite{2019GWNet}, on PEMS04 and PEMS08 datasets.
All referenced papers employ an identical experimental setup, \ie they utilize the last 12 time steps to predict the subsequent 12, and they report MAE, RMSE, and MAPE results for the prediction. 
Each row in the table thus presents performance findings for Graph WaveNet~(GWNet in short) or DCRNN as reported in experimental studies in different papers.

We can see a considerable performance variation for each solution across the different papers.
We also note that GWNet and DCRNN provide publicly available source code. As such, this variation is likely due to the varying training pipelines employed in the different studies.
Furthermore, our benchmark yields markedly improved performance compared to the results reported in the papers, with a maximum gap of 33\% (MAE of GWNet on PEMS04). 
To reduce spurious variations such as those just reported, we conduct a comprehensive analysis of existing codebases, and identify three primary sources of spurious variations: \textit{data processing}, \textit{training configurations}, and \textit{evaluation implementation}. These aspects are often overlooked, although they influence evaluation results substantially.

\begin{itemize}[leftmargin=*]
    \item \textit{Data Processing:} 
    A crucial step in the learning or inference process involves normalizing raw time series data. 
    Common approaches include min-max normalization and z-score normalization, each exerting varying effects on prediction performance.
    For example, some studies~\cite{2021ASTGNN} employ min-max normalization, whereas most studies usually adopt z-score normalization.
    \item \textit{Training Configurations:} Training configurations include optimization strategies and various training tricks. 
    Different setups have substantial impact on the optimization.
    For example, most studies~\cite{2018DCRNN,2019GWNet,2022STEP,2022D2STGNN} employ \textit{masked MAE} for model training, which excludes abnormal values that may affect predictions for normal values adversely.
    In contrast, some studies~\cite{STGODE, 2021Z-GCNets} adopt \textit{naive MAE} as their optimization function, which tends to yield inferior results.
    Further, the incorporation of training tricks, such as gradient clipping and curriculum learning, may also influence performance significantly~\cite{2022D2STGNN}. 
    \item \textit{Evaluation Implementation:} 
    While metrics have precise definitions, their implementations can vary across studies, including aspects such as handling outliers, and mini-batch computations~\cite{2021GTS}. 
    This difference results in significant deviations between testing and actual performance.
\end{itemize}

\begin{table}[t]
\renewcommand\arraystretch{1}
\setlength\tabcolsep{8pt}
    \centering
    \caption{Evaluation on normalized and re-normalized data. }
    \label{tab:unreasonanle_evaluation}
    \scalebox{0.75}{
    \begin{tabular}{cc|cc|ccr}
      \toprule
      \midrule
      \multicolumn{1}{c}{\multirow{2}*{\textbf{Data}}} & \multicolumn{1}{c}{\multirow{2}*{\textbf{Method}}} & \multicolumn{2}{c}{\textbf{normalized}} & \multicolumn{3}{c}{\textbf{re-normalized}}\\ 
      \cmidrule(r){3-4} \cmidrule(r){5-7} 
       & & MAE & MSE & MAE & MAPE & WAPE \\
      \hline
      \midrule
      \multirow{4}*{\rotatebox{90}{\textbf{ETTh1}}}
      & Autoformer~\cite{2021AutoFormer} &0.483 & 0.510 & 1.74 & {\cellcolor{pink}69.96\%} & {\cellcolor{pink}37.61\%} \\
      & FEDformer~\cite{2022FEDFormer} &0.460 & 0.467 & 1.71 & {\cellcolor{pink}68.92\%} & {\cellcolor{pink}36.89\%} \\
      & Crossformer~\cite{2023Crossformer} & 0.456 & 0.461 & 1.83 & {\cellcolor{pink}64.96\%} & {\cellcolor{pink}39.44\%} \\
      & PatchTST~\cite{2023PatchTST}  &0.426 & 0.432 &1.60 & {\cellcolor{pink}64.38\%} & {\cellcolor{pink}34.49\%} \\
      \hline
      \midrule
      \multirow{4}*{\rotatebox{90}{\textbf{ETTh2}}}
      & Autoformer~\cite{2021AutoFormer} &0.448 & 0.433 & 3.40 & {\cellcolor{pink}59.17\%} & {\cellcolor{pink}22.67\%} \\
      & FEDformer~\cite{2022FEDFormer} &0.431 & 0.418 & 3.35 & {\cellcolor{pink}56.14\%} & {\cellcolor{pink}22.33\%} \\
      & Crossformer~\cite{2023Crossformer} & 0.453 & 0.447 & 3.72 & {\cellcolor{pink}66.76\%} & {\cellcolor{pink}24.76\%} \\
      & PatchTST~\cite{2023PatchTST} &0.395 & 0.390 & 2.97 & {\cellcolor{pink}55.22\%} & {\cellcolor{pink}19.78\%} \\
      \midrule
      \bottomrule
    \end{tabular}
    }
  \end{table}

\subsubsection{Implementation of \BasicTS}
\BasicTS introduces a unified training pipeline, as depicted in Figure \ref{fig:BasicTS}. 
This mainly incorporates \textit{unified dataloader}, \textit{runner}, and \textit{evaluation} components to address the identified sources of spurious performance variations.
The \textit{unified dataloader} is equipped with z-score normalization as the default choice, which generally yields superior performance. 
Additionally, it adds external temporal features to the raw data such as time-of-day and day-of-week attributes.
The \textit{unified runner} controls the entire training, validation, and testing procedure. 
By default, we employ masked MAE as the loss function, which typically outperforms alternatives like naive MAE and MSE. 
Moreover, \textit{the unified runner} integrates commonly-used training tricks like curriculum learning and gradient clipping.
Lastly, the \textit{unified evaluation} component provides standard implementations of metrics including MAE, RMSE, MAPE, WAPE, MSE, and their masked versions. 
The three components form the foundation that enables \BasicTS  to support fair analyses and comparisons.
Given a model that conforms to the \textit{standard model interface}, \BasicTS can produce evaluation results for that model.
Furthermore, \BasicTS offers many \textit{extensibility features}, such as a logging system, customizable losses and metrics, and compatibility with diverse devices.

\begin{figure}[t]
  \centering
  \includegraphics[width=0.9\linewidth]{figures/BasicTS.pdf}
  \caption{Architecture of \BasicTS.}
  \label{fig:BasicTS}
\end{figure}

\subsection{Evaluation Settings}
Evaluation results should be presented in a clear and intuitive manner. 
In LTSF, many studies adopt metrics such as MAE and MSE and report the prediction performance based on \textit{normalized data} (z-score normalized). 
However, MAE and MSE represent absolute errors that can be influenced significantly by the range of the data, rendering them less intuitive for interpretation. 
Additionally, evaluating prediction performance on normalized data can yield seemingly very low prediction errors, potentially misleading readers unfamiliar with the details. 
Thus, some approaches to reporting prediction performance make it difficult for readers to judge whether the prediction performance of the model is satisfactory.

We suggest a practical approach: evaluating on re-normalized data and incorporating additional metrics such as MAPE and WAPE. 
The performance of important LTSF models on ETTh1 and ETTh2 datasets with normalization and re-normalization are summarized in Table \ref{tab:unreasonanle_evaluation}.
We can see that the prediction performance appears less satisfactory on the re-normalized data when considering the high MAPE and WAPE values, in contrast to the seemingly low MAE and MSE values obtained on the normalized data. 

{\color{black}
In summary, our evaluation is conducted on re-normalized data, employing metrics such as MAE, RMSE, MAPE, and WAPE.
Assuming $\Omega$ represents the indices of all observed samples, $y_i$ denotes the $i$-th actual sample, and $\hat{y_i}$ denotes the corresponding prediction, these metrics are defined as follows.

\begin{equation}
    \scalebox{0.84}{$
    \begin{aligned}
    \text{MAE}(y, \hat{y}) =& \frac{1}{|\Omega|} \sum_{i\in\Omega} |y_i - \hat{y}_i|, \quad
    \text{RMSE}(y, \hat{y}) =& \sqrt{\frac{1}{|\Omega|} \sum_{i\in\Omega} (y_i - \hat{y}_i)^2} \\
    \text{MAPE}(y, \hat{y}) =& \frac{1}{|\Omega|} \sum_{i\in\Omega} \left|\frac{y_i - \hat{y}_i}{y_i}\right|, \quad
    \text{WAPE}(y, \hat{y}) =& \frac{\sum_{i\in\Omega} |y_i - \hat{y}_i|}{\sum_{i\in\Omega} |y_i|}
    \end{aligned}$}
\end{equation}
The MAE and RMSE metrics quantify the prediction accuracy, while MAPE and WAPE serve to eliminate the influence of data units.
Additionally, for the M4 dataset, we adopt sMAPE, MASE, and OWA. For brevity, we omit their formulations and refer interested readers to the literature ~\cite{NBeats}.}

\section{Heterogeneity across MTS Datasets}

\label{sec:het_data}
Next, we put focus on the heterogeneity across MTS datasets and delve into its role in explaining the seemingly contradictory experimental findings that suggest that each of two different technical approaches is the best approach to achieve improved forecasting accuracy.
Unlike datasets in computer vision or natural language processing, which often share common patterns, MTS datasets can exhibit very distinct patterns derived from the diverse underlying systems. 
We thoroughly investigate this heterogeneity and categorize datasets based on characteristics of their temporal and spatial aspects. 
We argue that different types of patterns entail different solution challenges, implying that specific technical approaches are applicable \textit{only} to particular types of data. 
Neglecting this data heterogeneity can lead to seemingly conflicting experimental finding and to failure to advocate the right technical approach.

\subsection{Temporal Aspect}
\label{sec:temporal}

\begin{figure*}[t]
  \centering
  \includegraphics[width=0.9\linewidth]{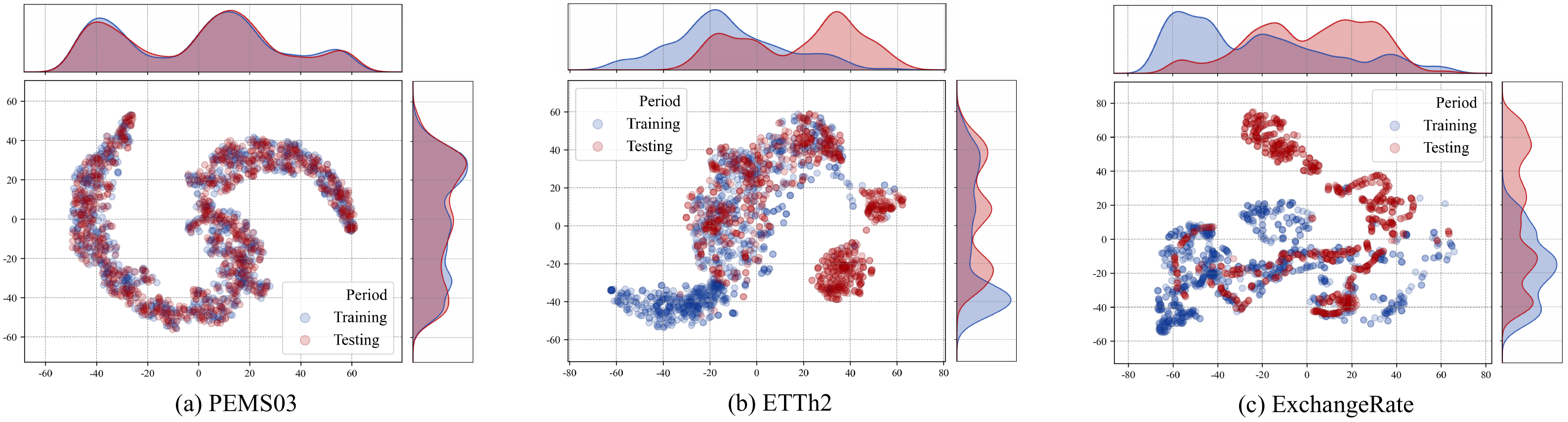}
  \caption{Visualization of data distribution based on t-SNE and kernel density estimation.}
  \label{fig:kde}
\end{figure*}

\begin{figure}[t]
  \centering
  \includegraphics[width=1\linewidth]{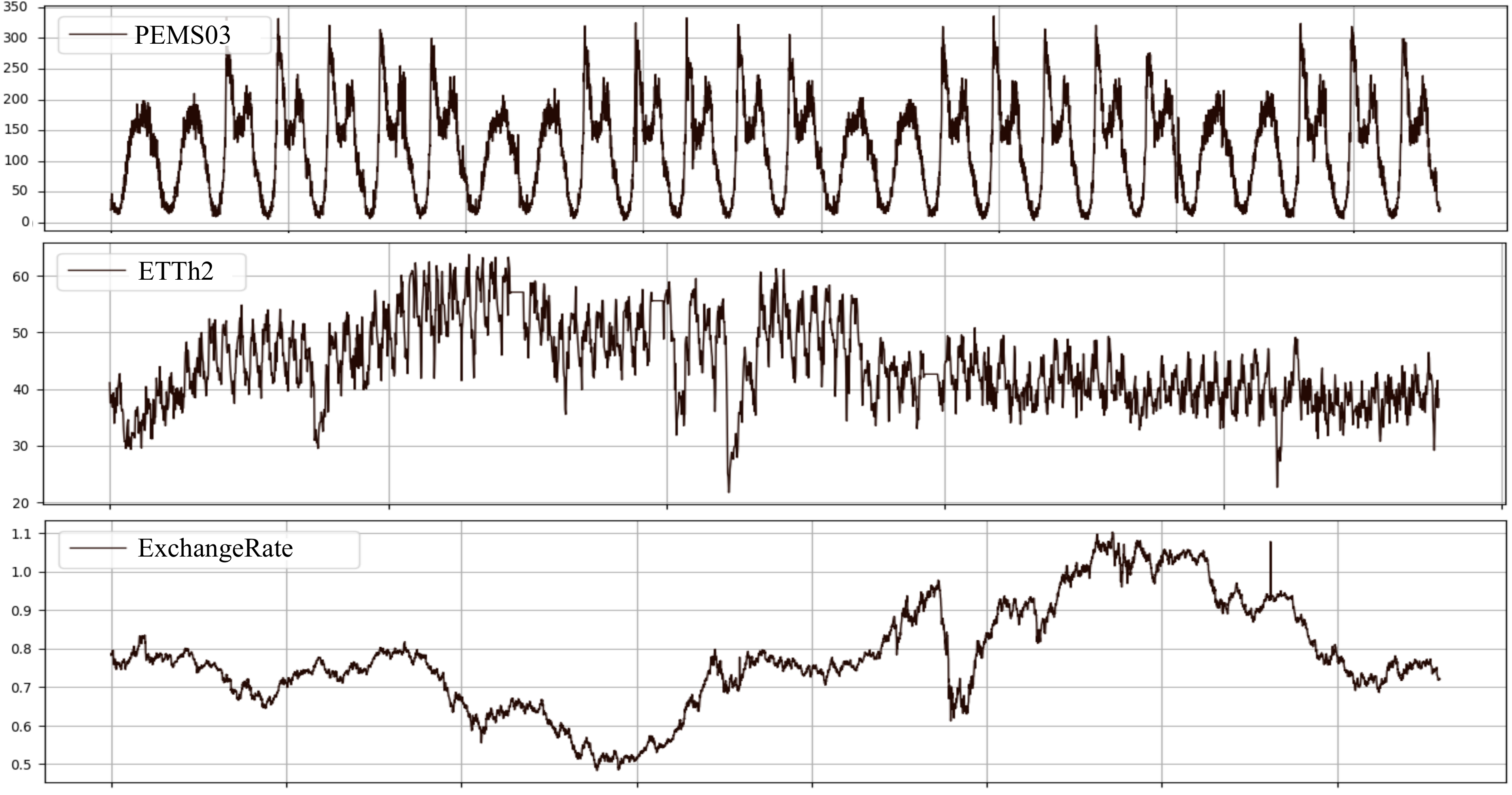}
  \caption{Distinct temporal patterns in multiple MTS datasets.}
  \label{fig:temporal_patterns}
\end{figure}

We categorize MTS datasets into three types according to their temporal aspect: datasets with clear and stable patterns, datasets with significant distribution drift, and datasets with unclear patterns.
We argue that these types of datasets are progressively less predictable.
However, quantifying the predictability~\cite{Predictability1,Predictability2} remains an unsolved challenge.
Thus, we analyze selected datasets through visualizations.
Specifically, we chose three typical datasets—PEMS03, ETTh2, and ExchangeRate—and visualized the original time series in Figure \ref{fig:temporal_patterns}.
{\color{black}To facilitate more intuitive comparisons, we reduce the dimensionality of these datasets to 2D using the t-SNE algorithm~\cite{tsne}, and then visualize the data distribution of the training and testing sets with the kernel density estimation algorithm~\cite{kde}, as shown in Figure \ref{fig:kde}.\footnote{The results shown in Figure \ref{fig:kde} are derived from time series datasets, where samples are obtained by sliding a window of size $P+F$ over the original time series (\ie the time series in Figure \ref{fig:temporal_patterns}). Here, $P$ and $F$ represent the lengths of the historical and future time series, respectively. For the PEMS03 dataset, $P$ and $F$ are set to 12, while for the ETTh2 and ExchangeRate datasets, they are set to 336. The selection of $P$ and $F$ is based on previous works~\cite{2022D2STGNN, 2023PatchTST}, and using different values for $P$ and $F$ yields similar results.}
}

We can see significantly distinct patterns across these datasets.
First, PEMS03, which records urban traffic flow at different locations, exhibits clear and stable patterns, \ie periodicity with a fixed period.
This pattern conforms to the overall periodicity and stability of urban traffic.
Second, ETT contains data from transformer sensors.
Although it contains evident cyclic patterns, the period is not fixed, and the mean is shifting, indicating distribution drift.
This is because the measured values are affected by external, unobserved factors, such as weather and sensor quality.
Third, ExchangeRate records the exchange rates of several currencies and displays minimally discernible patterns.
This outcome stems from the fact that exchange rates are primarily governed by unpredictable factors, such as economic policies.
Thus, historical data offers limited value for predictions, particularly for long-term predictions.
{\color{black}Additionally, as depicted in Figure \ref{fig:kde}, the data distributions of the training and testing sets in PEMS03 exhibit a high degree of similarity, whereas in the cases of ETTh2 and ExchangeRate, such similarity is not observed.}

{\color{black}Expanding on these insights, we argue that the inherent heterogeneity of MTS data is a key cause of seemingly conflicting findings when comparing advanced neural networks~\cite{2021Informer,2021AutoFormer,2022FEDFormer,2023PatchTST} and basic neural networks~\cite{2023DLinear}.
Advanced models usually possess strong data fitting capabilities.} When coupled with a strong inductive bias, this means that such models imply strong assumptions about data distributions. 
Conversely, due to their simplicity, {\color{black}basic models like the linear model~\cite{2023DLinear} struggle to capture complex patterns, but also feature relatively weak inductive bias.}
Considering both the different modeling capabilities of these approaches and the heterogeneous temporal patterns in MTS data
we argue that when used on datasets with stable and clear patterns, {\color{black}advanced models should be able to capture complex patterns such as periodicity, while basic linear models remain \textit{\textbf{under-fitted}} due to their limited capacities.}
In contrast, when used on datasets with significant distribution drift or unclear patterns,  {\color{black}advanced models are more likely to capture spurious features present only in the training dataset, thus facing \textbf{\textit{over-fitting}} problems.}
Based on this discussion, we formulate the following hypothesis:
\begin{tcolorbox}[boxrule = 0.3mm, colback=green_hyp!20, colframe=green_hyp!100, title=Hypothesis 1:, coltitle=black, fonttitle=\bfseries, lefttitle=0pt, left=2pt, right=3pt, top=3pt, bottom=3pt]
\begin{itemize}[leftmargin=15pt]
    \item[\textbf{1.1}]  {\color{black}Advanced neural networks outperform basic ones} on datasets with clear and stable temporal patterns, while basic neural networks suffer from \textbf{\textit{under-fitting}}.\\
    \vspace{-0.6cm}\\
    \item[\textbf{1.2}]  {\color{black}Basic neural networks generally outperform advanced ones} on datasets with significant distribution drift and datasets with unclear patterns, while  {\color{black}advanced neural networks} suffer from \textbf{\textit{over-fitting}}.
\end{itemize}
\end{tcolorbox}
\noindent 
The study that proposes LTSF-Linear~\cite{2023DLinear} ignores dataset heterogeneity, and conducts experiments on datasets without clear and stable patterns, leading to the biased conclusion that Transformer architectures are ineffective at MTS forecasting. 
We study this hypothesis experimentally in Section \ref{sec:effective_of_Transformers_and_Linears}.

\subsection{Spatial Aspect}
\label{sec:spatial}

Unlike easy-to-see temporal patterns, spatial dependencies are harder to grasp, and it is also more difficult to find clear metrics that allow to distinguish among datasets according to their spatial aspects.
Many studies interpret spatial patterns loosely as interactions between time series, and they model them using GCNs, without discussing in depth how to understand and quantify such patterns.
Fortunately, two recent studies, ST-Norm~\cite{2021STNorm} and STID~\cite{2022STID}, point out that the \textit{indistinguishability of samples in the spatial dimension} (spatial indistinguishability in short) gets to the essence of spatial dependencies.
In the following, we adopt this idea and, for the first time, design quantitative metrics to distinguish heterogeneous datasets according to their spatial aspect.
Specifically, we partition MTS datasets into two types: those with and those without significant spatial indistinguishability, and  then we discuss when and how to model spatial dependencies.

In MTS forecasting, samples are generated using a sliding window of size $T_p + T_f$, where $T_p$ and $T_f$ denote the lengths of the historical data and future data.
Spatial indistinguishability means that for a given time $t$, we can expect to generate many samples with similar historical data but different future data. 
Simple regression models (\eg using Multi-Layer Perceptions~(MLP), RNNs) cannot predict different future data based on similar historical data. Put differently, they cannot distinguish the historical samples~\cite{2022STID}.
Based on this concept, we propose the following quantitative metrics:

\begin{equation}
    \scalebox{0.75}{$
    \begin{aligned}
    r_1 = &\frac{\sum_{t, i, j}\mathbb{I}(\mathbf{A}_{t,i,j}^P>e_u \land \mathbf{A}_{t,i,j}^F < e_l)}{T\cdot N\cdot N}, \quad r_2 = \frac{\sum_{t,i,j}\mathbb{I}(\mathbf{A}_{t,i,j}^P > e_u\land\mathbf{A}_{t,i,j}^F < e_l)}{\sum_{t,i,j}\mathbb{I}(\mathbf{A}_{t,i,j}^P > e_u)}\\
    &\mathbf{A}^{P}_{t,i,j}=\frac{\mathbf{X}_{t-T_p:t}^i\cdot \mathbf{X}_{t-T_p:t}^j}{\Vert \mathbf{X}_{t-T_p:t}^i\Vert \Vert \mathbf{X}_{t-T_p:t}^j\Vert}, \qquad \mathbf{A}^{F}_{t,i,j}=\frac{\mathbf{X}_{t:t+T_f}^i\cdot \mathbf{X}_{t:t+T_f}^j}{\Vert \mathbf{X}_{t:t+T_f}^i\Vert \Vert \mathbf{X}_{t:t+T_f}^j\Vert}\\
\end{aligned}$
}
\end{equation}

\begin{figure}[t]
  \centering
  \includegraphics[width=1\linewidth]{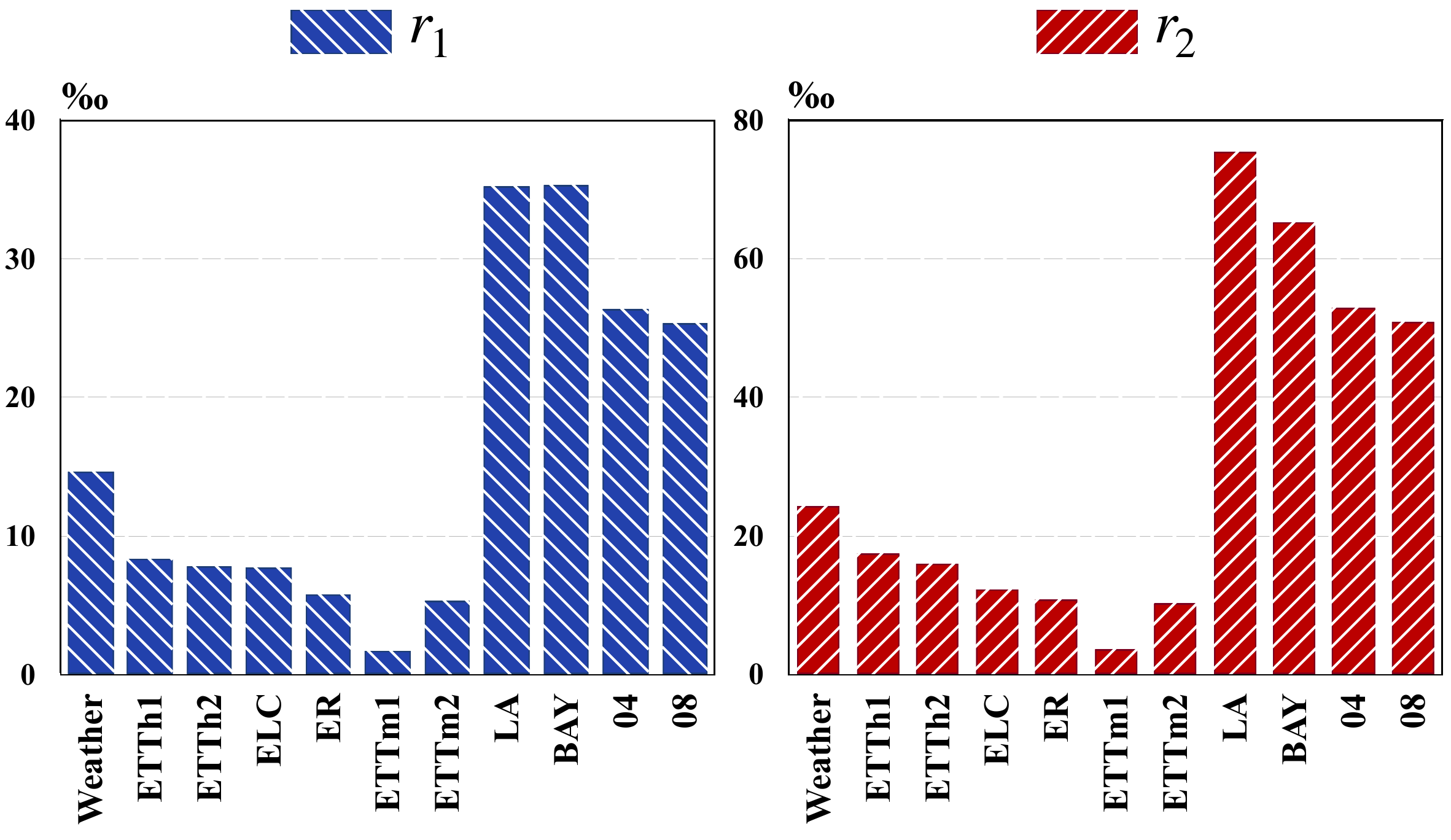}
  \caption{Spatial indistinguishability in different datasets.}
  \label{fig:spa_indisting}
\end{figure}

{\color{black}
\noindent \textbf{Intuitive Understanding}. For a dataset with $T$ time steps and $N$ samples, we construct two similarity matrices, $\mathbf{A}^P, \mathbf{A}^F\in\mathbb{R}^{T\times N\times N}$, representing pairwise similarities among the samples at each time step. Specifically, $\mathbf{A}^{*}_{t, i, j}$ denotes the similarity between time series $i$ and $j$ at time step $t$.
Using these matrices, we define the total sample count as $T\cdot N\cdot N$, the count of historically similar samples as $\sum_{i,j,t}\mathbb{I}(\mathbf{A}^P_{i,j,t}>e_u)$, and the count of indistinguishable samples as $\sum_{i,j,t}\mathbb{I}(\mathbf{A}^P_{i,j,t}>e_u\land\mathbf{A}^F_{i,j,t} < e_l)$. Here, $e_u = 0.9$ and $e_l = 0.5$ are the upper and lower similarity thresholds, respectively. The indicator function $\mathbb{I}(\cdot)$ returns 1 when the condition is satisfied, and 0 otherwise.
We then define two metrics: $r_1$, the ratio of indistinguishable samples to the total number of samples, and $r_2$, the ratio of indistinguishable samples to those with similar historical data. These metrics provide complementary insights: $r_1$ helps determine whether indistinguishability is a major obstacle to improving predictive performance, while $r_2$ offers a more nuanced evaluation of the degree of indistinguishability.
}

We calculate the above two metrics for 11 common datasets. 
The results are shown in Figure \ref{fig:spa_indisting}.
We can clearly see that the $r_1$ and $r_2$ of ETT, Electricity~(ELC), ExchangeRate~(ER), and Weather are very low, while the $r_1$ and $r_2$ of METR-LA~(LA), PEMS-BAY~(BAY), PEMS04~(04), and PEMS08~(08) are substantially higher.
Interestingly, although these two different groups of datasets have exactly the same format, they are rarely combined in experimental studies. ETT, Electricity, ExchangeRate, and Weather are often used in LTSF studies, where spatial dependencies are not of prime interest. Further, METR-LA, PEMS-BAY, PEMS04, and PEMS08 are used in STF studies, where spatial dependencies take center stage.

Given the above insights, we discuss when and how to model spatial dependencies. 
First, there is no urgent need to model spatial dependencies on datasets without significant spatial indistinguishability, and forcibly modeling spatial dependencies may even degrade performance. 
Second, on datasets with significant spatial indistinguishability, modeling spatial dependencies by addressing spatial indistinguishability can improve performance.
To be more specific, we discuss how STGNNs~\cite{2019GWNet, 2018DCRNN}, ST-Norm~\cite{2021STNorm}, and STID~\cite{2022STID} work.
First, GCNs in STGNNs~\cite{2019GWNet, 2018DCRNN} usually rely on graph structures that conform to the homophily assumption~\cite{homophily1, homophily2}, where connected nodes often share similar labels\footnote{In regression, the label is a real-value response corresponding to the instance~\cite{FPP}.}.
Therefore, nodes~(\ie time series) with similar historical data but different future data~(\ie labels) are often disconnected.
Given such graph structures, GCNs perform message aggregation to make historical data distinguishable.
Second, ST-Norm~\cite{2021STNorm} normalizes data on the spatial dimension by separately refining the high-frequency and the local components underlying the input data, making the historical data distinguishable as well.
Third, STID~\cite{2022STID} proposes a simple yet effective idea of attaching a trainable spatial identity to each time series to distinguish similar historical data.
Based on the above discussion, we state the following hypothesis:

\begin{tcolorbox}[boxrule = 0.3mm, colback=green_hyp!20, colframe=green_hyp!100, title=Hypothesis 2:, coltitle=black, fonttitle=\bfseries, lefttitle=0pt, left=2pt, right=3pt, top=3pt, bottom=3pt
]
\begin{itemize}[leftmargin=15pt]
    \item[\textbf{2.1}] On datasets with significant spatial indistinguishability, modeling spatial dependencies by addressing spatial indistinguishability can \textbf{improve} performance.\\
    \vspace{-0.6cm}\\
    \item[\textbf{2.2}] On datasets without significant spatial indistinguishability, forcing the modeling towards spatial dependencies may \textbf{degrade} performance degradation.
\end{itemize}
\end{tcolorbox}
\noindent We study this hypothesis in Section \ref{sec:when_and_how_to_model_spatial_dependencies}.

\section{Experiments}
\label{sec:experiments}

In this section, we conduct extensive experiments to assess our hypotheses and address controversies in technical approaches.
In addition, we provide comprehensive analysis and comparison of popular MTS forecasting models based on \BasicTS and offer insight into the progress already made.
Specifically, Section~\ref{sec:exp_setup} covers datasets, baselines, and implementation details. 
{\color{black}
Section~\ref{sec:effective_of_Transformers_and_Linears} evaluates the effectiveness of advanced and basic neural networks for LTSF, thus confirming the hypothesis presented in Section V-A~\ref{sec:temporal}.
}
Section \ref{sec:when_and_how_to_model_spatial_dependencies}  consider when and how to model spatial dependencies, confirming the hypothesis in Section \ref{sec:spatial}. 
{\color{black}
Section \ref{sec:main_results} discusses how to select models or datasets, presents detailed experimental results, and offers insight into the advancements made.}
All code, datasets, experimental scripts, and results can be accessed through the public GitHub repository at {\color{blue}\url{https://github.com/GestaltCogTeam/BasicTS}}.

\subsection{Experimental Setup}
\label{sec:exp_setup}
\subsubsection{Datasets}
Following previous LTSF and STF studies~\cite{2018LSTNet, 2021Informer, 2021AutoFormer, 2019ASTGCN, 2018DCRNN}, we use 14 datasets to conduct experiments, including METR-LA, PEMS-BAY, PEMS03, PEMS04, PEMS07, PEMS08, ETTh1, ETTh2, ETTm1, ETTm2, Electricity, Weather, ExchangeRate, and M4 datasets. 
Not all the datasets from \BasicTS are included due to space limitations.
The remaining datasets are available via the code repository, including large-scale MTS datasets~\cite{2023LargeST}.

\subsubsection{Baselines}
{\color{black}
We include popular baselines for which official code is available, including LTSF and STF models. 
For brevity, we omit their detailed descriptions and simply categorize the baselines based on their technical approaches.

Considering STF models, we cover influential baselines that have high citation counts or offer state-of-the-art performance. 
First, STGCN~\cite{2018STGCN}, DCRNN~\cite{2018DCRNN}, GWNet~\cite{2019GWNet}, DGCRN~\cite{2021DGCRN}, and D$^2$STGNN \cite{2022D2STGNN} are prior-graph-based solutions that rely on pre-defined graphs to indicate spatial dependencies among time series.
Second, AGCRN~\cite{2020AGCRN}, MTGNN~\cite{2020MTGNN}, StemGNN~\cite{2020StemGNN}, GTS~\cite{2021GTS}, and STEP~\cite{2022STEP} are latent-graph-based methods that learn graph structures and optimize STGNNs jointly.
Third, we adopt two non-graph based methods, ST-Norm~\cite{2021STNorm} and STID~\cite{2022STID}.

Considering LTSF models, we cover both {\color{black}advanced and basic neural networks.}
First, Informer~\cite{2021Informer}, Autoformer~\cite{2021AutoFormer}, FEDformer~\cite{2022FEDFormer}, Triformer~\cite{Triformer}, Pyraformer~\cite{2022Pyraformer}, Crossformer~\cite{2023Crossformer}, PatchTST~\cite{2023PatchTST} utilize variants of the Transformer to capture long-term historical information.
Second, Linear, DLinear, and NLinear utilize a simple linear layer~\cite{2023DLinear}.

{\color{black}
For a more exhaustive comparison, we also cover three classic time series forecasting models: LGBM~\cite{LightGBM}, DeepAR~\cite{DeepAR}, and NBeats~\cite{NBeats}. LGBM is a widely-used gradient boosting framework. DeepAR~\cite{DeepAR} and NBeats~\cite{NBeats} are classic deep learning solutions.
These baselines are adopted widely in many industrial applications.}

Due to the space limitation, we cannot cover all baselines in \BasicTS; additional baselines are included in the repository, \eg STGODE~\cite{STGODE}, NHiTS~\cite{NHiTS}, and TimesNet~\cite{2022TimesNet}.


\begin{table*}[htbp]
\setlength\tabcolsep{4pt}
\renewcommand\arraystretch{0.8}
    \centering
    \caption{\color{black}Performance of advanced Transformer models and basic linear models across heterogeneous MTS datasets. }
    \label{tab:effectiveness_of_transformers}
    \scalebox{0.8}{
    \begin{tabular}{c|ccr|ccr|ccr|ccr}
      \toprule
      \midrule
      \multicolumn{1}{c}{\multirow{2}*{\textbf{Methods}}} &\multicolumn{3}{c}{\textbf{PEMS04}} &\multicolumn{3}{c}{\textbf{PEMS08}} & \multicolumn{3}{c}{\textbf{ETTh2}} & \multicolumn{3}{c}{\textbf{ETTm2}}\\ 
      \cmidrule(r){2-4} \cmidrule(r){5-7} \cmidrule(r){8-10} \cmidrule(r){11-13}
       & MAE & RMSE & WAPE & MAE & RMSE & WAPE & MAE & RMSE & WAPE & MAE & RMSE & WAPE\\
      \midrule
     Informer& {\cellcolor{green_bg}{27.94}} & {\cellcolor{green_bg}{44.74}} & {\cellcolor{green_bg}{12.84\%}} & {\cellcolor{green_bg}{26.92}} & {\cellcolor{green_bg}{43.79}} & {\cellcolor{green_bg}{11.63\%}} & 7.12 & 6.87 & 47.44\% & 5.84 & 7.90 & 38.97\%\\
     Autoformer & {\cellcolor{green_bg}34.72} & {\cellcolor{green_bg}50.33} & {\cellcolor{green_bg}14.81\%} & {\cellcolor{green_bg}33.75} &{\cellcolor{green_bg}51.23} & {\cellcolor{green_bg}14.13\%} & 3.33 & \textbf{4.91} & 22.17\% & 2.74 & 4.58 & 18.27\%\\
     FEDformer & {\cellcolor{green_bg}\textbf{26.89}} & {\cellcolor{green_bg}\textbf{41.46}} & {\cellcolor{green_bg}\textbf{12.39\%}} & {\cellcolor{green_bg}\textbf{25.14}} &{\cellcolor{green_bg} \textbf{39.17}} & {\cellcolor{green_bg}\textbf{10.87\%}} & \textbf{3.27} & {4.93} & \textbf{21.78\%} & \textbf{2.70} & \textbf{4.54} & \textbf{17.99\%}\\
      \midrule
    Linear & 37.42 & \textbf{62.14} & 17.22\% & \textbf{34.04} & \textbf{57.07} & \textbf{14.71\%} & {\cellcolor{green_bg}3.18} & {\cellcolor{green_bg}5.04} & {\cellcolor{green_bg}21.19\%} & {\cellcolor{green_bg}2.52} & {\cellcolor{green_bg}4.24} & {\cellcolor{green_bg}16.80\%}\\
    DLinear & \textbf{37.51} & 62.21 & \textbf{17.26\%} & 34.15 & 57.18 & 14.76\% & {\cellcolor{green_bg}{\color{black}\textbf{3.13}}} & {\cellcolor{green_bg}\textbf{5.00}} & {\cellcolor{green_bg}{\color{black}\textbf{20.85\%}}} & {\cellcolor{green_bg}\textbf{2.49}} & {\cellcolor{green_bg}4.23} & {\cellcolor{green_bg}{16.63\%}}\\
    NLinear & 37.62 & 62.38 & 17.31\% & 34.11 & 57.26 & 14.74\% & {\cellcolor{green_bg}3.16} & {\cellcolor{green_bg}5.06} & {\cellcolor{green_bg}21.09\%} & {\cellcolor{green_bg}\textbf{2.49}} & {\cellcolor{green_bg}\textbf{4.21}} & {\cellcolor{green_bg}\textbf{16.60\%}}\\
      \midrule
      Gap & {\color{red}\textbf{39.49\%}$\downarrow$} & {\color{red}\textbf{49.87\%}$\downarrow$} & {\color{red}\textbf{39.30\%}$\downarrow$} & {\color{red}\textbf{35.40\%}$\downarrow$} & {\color{red}\textbf{45.69\%}$\downarrow$} & {\color{red}\textbf{35.32\%}$\downarrow$} & {\color{teal}\textbf{4.28\%}$\uparrow$} & {\color{teal}\textbf{1.83\%}$\uparrow$} & {\color{teal}\textbf{4.26\%}$\uparrow$} & {\color{teal}\textbf{7.78\%}$\uparrow$} & {\color{teal}\textbf{7.27\%}$\uparrow$} & {\color{teal}\textbf{7.72\%}$\uparrow$}\\
      \midrule
      \bottomrule
    \end{tabular}
    }
  \end{table*}

\subsubsection{Implementation details}
\label{sec:implementation_details}
For dataset partitioning, we adopt settings consistent with previous work for each dataset. For brevity, we omit the details and refer interested readers to our repository. 
We set the length of the historical data and future data of the STF task to 12. 
For the LTSF task, we set the length of future data to 336. We vary the historical length among 96, 192, 336, and 720, and we report the best prediction performance.
For error calculations, we report only the \textit{average} error between the forecast time series and the true future time series, due to the space limitation.
For the STF task, we employ the MAE, RMSE, MAPE, and WAPE metrics. 
For the LTSF task, we disregard MAPE, considering that there are many zero values in commonly used LTSF datasets.
{\color{black}For the M4 competition dataset, we employ its original settings~\cite{M4Paper}.}
In the efficiency studies in Section \ref{sec:main_results}, we report the average training time per epoch (in seconds) and the number of model parameter (in million).
We set the batch size to 64. If an Out-Of-Memory~(OOM) situation occurs, we reduce the batch size by half (to a minimum of 8).
All experiments are conducted using a NVIDIA 3090 GPU and 128 GB memory.

{\color{black}
\subsubsection{Hyperparameter tuning}
For model implementation, we adopt the public model architecture and hyperparameters.
For optimization hyperparameters, such as learning rate and batch size, we also adopt the public settings. Then, we tune these hyperparameters of each model on each dataset via grid search to ensure performance \textit{at least as good as reported in the original paper} (if available).
Although using AutoML to tune these hyperparameters may be optimal, we found that manual hyperparameter tuning is acceptable within a certain range. For example, batch sizes of 32, 64, and 128 yield similar performance and do not contradict our findings.}

\begin{figure}[t]
  \centering
  \includegraphics[width=0.9\linewidth]{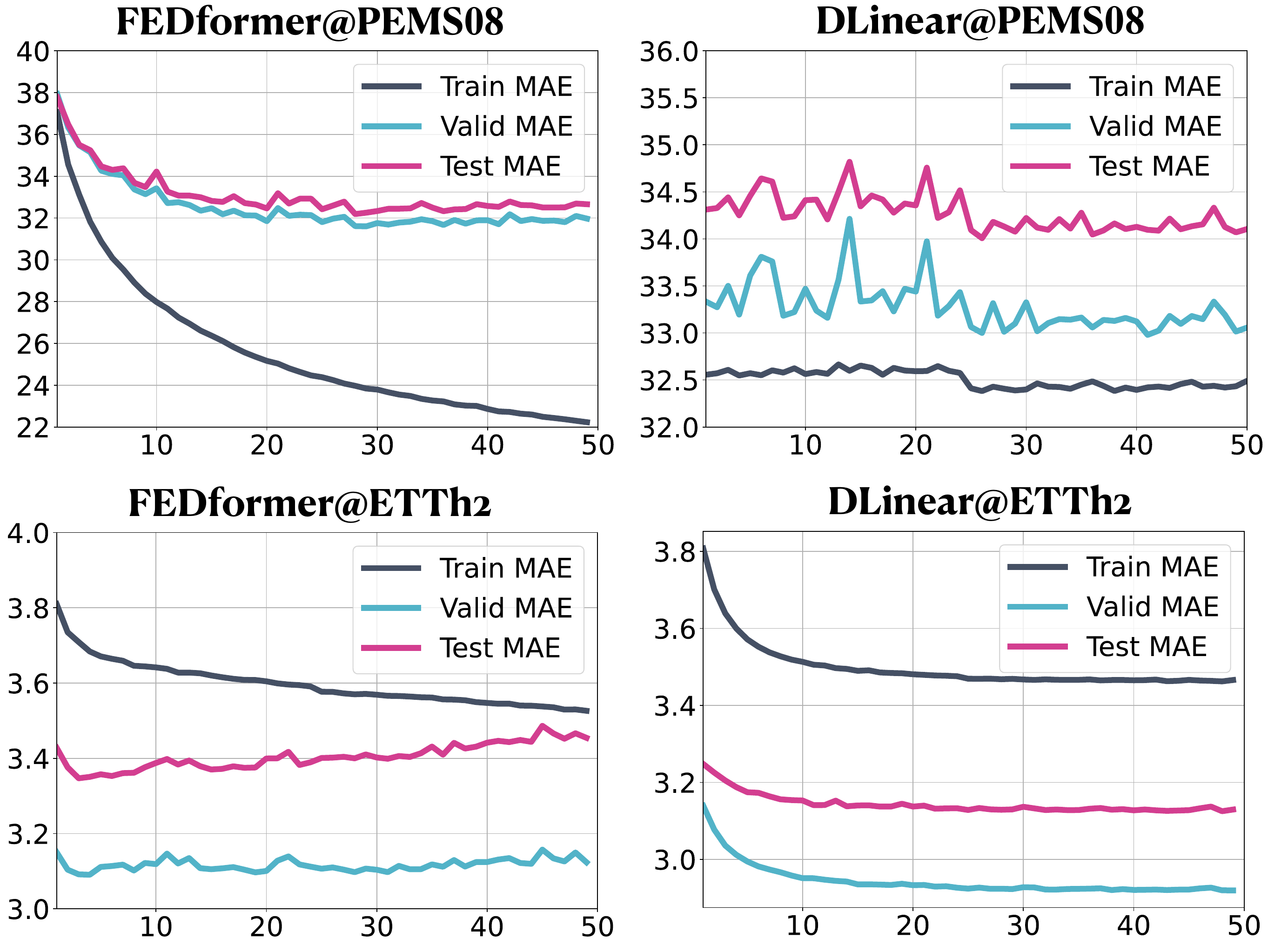}
  \caption{MAE for varying epochs.}
  \label{fig:under_fitting_over_fitting}
\end{figure}

{\color{black}\subsection{Advanced Neural Networks \textit{vs.} Basic Neural Networks}}
\label{sec:effective_of_Transformers_and_Linears}

{\color{black}This subsection studies the performance of advanced models~(\eg Transformers) versus basic models~(\eg linear models) and assesses the hypotheses in Section~\ref{sec:temporal}.}
We consider four datasets: 
PEMS04 and PEMS08, which exhibit clear and stable patterns, and ETTh2 and ETTm2, which demonstrate significant distribution drift or unclear patterns.
%
%
{\color{black}
Six baseline models are chosen based on the LTSF-Linear study~\cite{2023DLinear}, where Informer, Autoformer, and FEDformer are advanced Transformer models, and Linear, DLinear, and NLinear are basic linear models.}
They all follow the LTSF setup described in Section \ref{sec:implementation_details}. We report MAE, RMSE, and WAPE.
{\color{black}Furthermore, we calculate the performance gap between the best advanced and basic models, as shown in Table \ref{tab:effectiveness_of_transformers}}.

 {\color{black}First, advanced models generally outperform basic models by a very large margin~(green background) on datasets with clear and stable patterns.
Second, basic models consistently outperform advanced models on datasets with distribution drifts or unclear patterns.}
This gap in prediction performance may at first seem puzzling.
To intuitively understand why, we visualize the MAE when varying the number of epochs for FEDformer and DLinear on PEMS08 and ETTh2 datasets---see Figure \ref{fig:under_fitting_over_fitting}.
On PEMS08, the training, validation, and testing MAEs of FEDformer start from similar values and keep decreasing. 
In contrast, DLinear's MAEs, even the training MAE, do not decrease with increasing epochs, which indicates that DLinear suffers from \textbf{\textit{under-fitting}}.
Next, on the ETTh2 dataset, the training MAE of FEDformer keeps decreasing, while its validation and testing MAEs start to increase already when reaching 2 epochs, which indicates that FEDformer suffers from serious \textbf{\textit{over-fitting}}.
These results are consistent with the hypothesis in Section \ref{sec:temporal}.

We summarize our findings as follows. 
{\color{black}First, benefiting from their strong modeling capacities, advanced neural networks are far more effective than basic neural networks when the data has clear and stable patterns.}
Second, models with less inductive bias~\cite{2021ViT}~(\eg models based on MLPs or a vanilla Transformer~\cite{2023PatchTST}) usually perform better when there is no explicit pattern.
Moreover, although some recent solutions are positioned as general MTS prediction solutions, we believe that effective general solutions should first perform well on data with clear patterns and should then also consider their performance on time series with less clear patterns.

\begin{table}[b]
\setlength\tabcolsep{6pt}
\renewcommand\arraystretch{0.8}
    \centering
    \caption{Performance of STID, AGCRN, and their variants on datasets with varying spatial indistinguishability.}
    \label{tab:stid_agcrn}
    \scalebox{0.8}{

    \begin{tabular}{cc|cc|cc|c}
      \toprule
      \midrule
      {\textbf{Data}} & {\textbf{Metrics}} & {\textbf{STID}} & {\textbf{AGCRN}} & {\textbf{STID}$^*$} & {\textbf{AGCRN}$^*$} & {\textbf{Gap}} \\
      \midrule
      \multirow{3}*{{\rotatebox{90}{\textbf{LA}}}} & MAE & {\cellcolor{green_bg}{3.12}} & {\cellcolor{green_bg}3.16} & 3.58 & 3.36 & {\color{teal}\textbf{12.85}\%$\uparrow$}\\
      & RMSE & {\cellcolor{green_bg}{6.49}} & {\cellcolor{green_bg}6.44} & 7.24 & 6.90& {\color{teal}\textbf{10.35}\%$\uparrow$}\\
      & MAPE & {\cellcolor{green_bg}{9.13\%}} & {\cellcolor{green_bg}8.88\%} & 10.32\% & 9.66\% & {\color{teal}\textbf{11.53}\%$\uparrow$}\\
      \midrule
      \multirow{3}*{{\rotatebox{90}{\textbf{BAY}}}} & MAE & {\cellcolor{green_bg}{1.56}} & {\cellcolor{green_bg}1.60} & 1.80 & 1.70 & {\color{teal}\textbf{13.33}\%$\uparrow$}\\
      & RMSE & {\cellcolor{green_bg}{3.59}} & {\cellcolor{green_bg}3.67} & 4.21 & 3.96& {\color{teal}\textbf{14.72}\%$\uparrow$}\\
      & MAPE & {\cellcolor{green_bg}{3.53\%}} & {\cellcolor{green_bg}3.65\%} & 4.12\% & 3.92\%& {\color{teal}\textbf{14.32}\%$\uparrow$}\\
      \midrule
      \multirow{3}*{{\rotatebox{90}{\textbf{ER}}}} & MAE & 0.0325 & 0.0455 & {\cellcolor{green_bg}{0.0312}} & {\cellcolor{green_bg}0.0421} & {\color{red}\textbf{8.07}\%$\downarrow$} \\
      & WAPE & 4.28\% & 5.98\% & {\cellcolor{green_bg}{4.11\%}} & {\cellcolor{green_bg}5.51\%} & {\color{red}\textbf{8.52}\%$\downarrow$}\\
      & MAPE & 7.21\% & 12.03\% & {\cellcolor{green_bg}{6.89\%}} & {\cellcolor{green_bg}9.60\%} & {\color{red}\textbf{25.31}\%$\downarrow$}\\
      \midrule
      \multirow{3}*{{\rotatebox{90}{\textbf{\small ETTm1}}}} & MAE & 1.63 & 2.29 & {\cellcolor{green_bg}{1.41}} & {\cellcolor{green_bg}{1.89}} & {\color{red}\textbf{21.16}\%$\downarrow$}\\
      & WAPE & 35.24\% & 49.42\% & {\cellcolor{green_bg}{30.64\%}} & {\cellcolor{green_bg}{40.89\%}} & {\color{red}\textbf{20.86}\%$\downarrow$} \\
      & MAPE & 64.86\% & 75.64\% & {\cellcolor{green_bg}{55.13\%}} & {\cellcolor{green_bg}{68.39\%}} & {\color{red}\textbf{17.65}\%$\downarrow$}\\

      \midrule
      \bottomrule
    \end{tabular}
    }
  \end{table}

\begin{table*}[htpb]
\setlength\tabcolsep{4.3pt}
\renewcommand\arraystretch{0.60}
    \centering
    \caption{STF on METR-LA, PEMS-BAY, PEMS03, PEMS04, PEMS07, and PEMS08 datasets. }
    \label{tab:main_stf}
    \scalebox{0.7}{
    \begin{tabular}{cc|ccc|ccccc|ccccc|cc}
\toprule
\midrule
      \textbf{Data} & \textbf{Metrics} & \textbf{LGBM} & \textbf{DeepAR} & \textbf{NBeats} & \textbf{STGCN} & \textbf{DCRNN} & \textbf{GWNet} & \textbf{DGCRN} & \textbf{D$^2$STGNN} & \textbf{AGCRN} & \textbf{MTGNN} & \textbf{StemGNN} & \textbf{GTS} & \textbf{STEP} & \textbf{STNorm} & \textbf{STID}\\
\midrule
\multirow{6}*{\rotatebox{90}{\textbf{METR-LA\ }}}  & MAE & 5.03 & \underline{\textbf{3.33}} & 3.79 & 3.11 & 3.03 & 3.03 & 2.94 & \color{teal}{\underline{\textbf{2.88}}} & 3.16 & 3.05 & 3.72 & 3.13 & \underline{\textbf{2.93}} & 3.14 & \underline{\textbf{3.12}} \\
 & RMSE & 9.67 & \underline{\textbf{6.75}} & 7.74 & 6.31 & 6.23 & 6.12 & 6.04 & \color{teal}{\underline{\textbf{5.91}}} & 6.44 & 6.16 & 7.33 & 6.32 & \underline{\textbf{5.96}} & \underline{\textbf{6.49}} & \underline{\textbf{6.49}} \\
 & MAPE & 13.12\% & \underline{\textbf{9.76\%}} & 10.69\% & 8.37\% & 8.31\% & 8.17\% & \color{teal}{\underline{\textbf{7.79\%}}} & 7.83\% & 8.88\% & 8.16\% & 10.43\% & 8.62\% & \underline{\textbf{8.00\%}} & \underline{\textbf{8.84\%}} & 9.13\% \\
 & WAPE & 8.72\% & \underline{\textbf{5.76\%}} & 6.65\% & 5.38\% & 5.24\% & 5.24\% & 5.10\% & \color{teal}{\underline{\textbf{4.99\%}}} & 5.48\% & 5.28\% & 6.45\% & 5.42\% & \underline{\textbf{5.07\%}} & 5.44\% & \underline{\textbf{5.40\%}} \\
 \cmidrule{2-17}
 & Param & - & \underline{\color{teal}{\textbf{0.10}}} & 8.07 & 0.25 & 0.37 & 0.31 & \underline{\textbf{0.20}} & 0.39 & 0.75 & \underline{\textbf{0.41}} & 1.20 & 38.49 & 40.48 & 0.22 & {\underline{\textbf{0.12}}} \\
 & Speed & - & 24.48 & \underline{\textbf{11.36}} & \underline{\textbf{21.01}} & 94.87 & 27.70 & 128.84 & 152.33 & 28.22 & 24.37 & \underline{\textbf{16.19}} & 52.23 & 497.26 & 25.26 & \color{teal}{\underline{\textbf{7.50}}} \\
\midrule
\multirow{6}*{\rotatebox{90}{\textbf{PEMS-BAY\ }}}  & MAE & 2.10 & \underline{\textbf{1.70}} & 1.95 & 1.63 & 1.59 & 1.59 & 1.58 & \underline{\textbf{1.52}} & 1.60 & 1.60 & 1.99 & 1.68 & \color{teal}{\underline{\textbf{1.48}}} & 1.58 & \underline{\textbf{1.56}} \\
 & RMSE & 4.63 & \underline{\textbf{3.84}} & 4.96 & 3.72 & 3.69 & 3.68 & 3.65 & \underline{\textbf{3.53}} & 3.67 & 3.71 & 4.49 & 3.79 & \color{teal}{\underline{\textbf{3.42}}} & 3.65 & \underline{\textbf{3.59}} \\
 & MAPE & 4.98\% & \underline{\textbf{3.83\%}} & 4.43\% & 3.69\% & 3.58\% & 3.60\% & 3.52\% & \underline{\textbf{3.44\%}} & 3.65\% & 3.59\% & 4.61\% & 3.78\% & \color{teal}{\underline{\textbf{3.31\%}}} & \underline{\textbf{3.52\%}} & 3.53\% \\
 & WAPE & 3.37\% & \underline{\textbf{2.72\%}} & 3.13\% & 2.61\% & 2.55\% & 2.55\% & 2.53\% & \underline{\textbf{2.43\%}} & 2.56\% & 2.57\% & 3.19\% & 2.68\% & \color{teal}{\underline{\textbf{2.37\%}}} & 2.52\% & \underline{\textbf{2.50\%}} \\
 \cmidrule{2-17}
 & Param & - & \underline{\color{teal}{\textbf{0.11}}} & 8.07 & 0.31 & 0.37 & 0.31 & \underline{\textbf{0.21}} & 0.40 & 0.75 & \underline{\textbf{0.57}} & 1.39 & 58.67 & 60.47 & 0.28 & {\underline{\textbf{0.12}}} \\
 & Speed & - & \underline{\textbf{42.79}} & 45.24 & \underline{\textbf{51.91}} & 223.35 & 70.24 & 401.66 & 374.32 & 62.07 & 46.47 & \underline{\textbf{30.30}} & 179.98 & 1406.90 & 63.90 & \color{teal}{\underline{\textbf{13.28}}} \\
\midrule
\multirow{6}*{\rotatebox{90}{\textbf{PEMS03\ \ }}}  & MAE & 20.56 & \underline{\textbf{16.63}} & 19.71 & 15.83 & 15.54 & \color{teal}{\underline{\textbf{14.59}}} & 14.60 & 14.63 & 15.24 & \underline{\textbf{14.85}} & 16.95 & 15.41 & N/A & \underline{\textbf{15.32}} & 15.33 \\
 & RMSE & 34.19 & \underline{\textbf{28.36}} & 32.52 & 27.51 & 27.18 & \underline{\textbf{25.24}} & 26.20 & 26.31 & 26.65 & \color{teal}{\underline{\textbf{25.23}}} & 28.52 & 26.15 & N/A & \underline{\textbf{25.93}} & 27.40 \\
 & MAPE & 22.58\% & \underline{\textbf{17.76\%}} & 19.27\% & 16.13\% & 15.62\% & 15.52\% & \underline{\textbf{14.87\%}} & 15.32\% & 15.89\% & \underline{\textbf{14.55\%}} & 19.61\% & 15.39\% & N/A & \color{teal}{\underline{\textbf{14.37\%}}} & 16.40\% \\
 & WAPE & 11.82\% & \underline{\textbf{9.56\%}} & 11.34\% & 9.11\% & 8.94\% & \color{teal}{\underline{\textbf{8.39\%}}} & 8.40\% & 8.42\% & 8.77\% & \underline{\textbf{8.54\%}} & 9.75\% & 8.86\% & N/A & \underline{\textbf{8.81\%}} & 8.82\% \\
 \cmidrule{2-17}
 & Param & - & \underline{\color{teal}{\textbf{0.11}}} & 8.07 & 0.32 & 0.37 & 0.31 & \underline{\textbf{0.21}} & 0.40 & 0.75 & \underline{\textbf{0.62}} & 1.46 & 25.27 & N/A & 0.30 & {\underline{\textbf{0.12}}} \\
 & Speed & - & \underline{\textbf{20.41}} & 20.60 & \underline{\textbf{25.55}} & 102.77 & 30.82 & 191.13 & 187.48 & 28.92 & 21.66 & \underline{\textbf{13.50}} & 59.16 & N/A & 29.58 & \color{teal}{\underline{\textbf{5.78}}} \\
\midrule
\multirow{6}*{\rotatebox{90}{\textbf{PEMS04\ \ }}}  & MAE & 26.56 & \underline{\textbf{20.64}} & 25.30 & 19.76 & 19.66 & 18.80 & 18.84 & \color{teal}{\underline{\textbf{18.32}}} & 19.28 & 19.13 & 22.98 & 21.32 & \color{teal}{\underline{\textbf{18.32}}} & 19.21 & \underline{\textbf{18.35}} \\
 & RMSE & 41.61 & \underline{\textbf{32.35}} & 39.65 & 31.51 & 31.18 & 30.14 & 30.48 & \underline{\textbf{29.89}} & 31.02 & 31.03 & 36.00 & 33.55 & \underline{\textbf{29.91}} & 32.30 & \color{teal}{\underline{\textbf{29.86}}} \\
 & MAPE & 18.96\% & \underline{\textbf{14.28\%}} & 17.66\% & 13.48\% & 13.45\% & 13.19\% & 12.92\% & \underline{\textbf{12.51\%}} & 13.18\% & 13.22\% & 16.56\% & 14.85\% & \underline{\textbf{12.60\%}} & 13.05\% & \color{teal}{\underline{\textbf{12.50\%}}} \\
 & WAPE & 12.09\% & \underline{\textbf{9.38\%}} & 11.51\% & 8.98\% & 8.94\% & 8.55\% & 8.57\% & \color{teal}{\underline{\textbf{8.33\%}}} & 8.77\% & 8.70\% & 10.45\% & 9.70\% & \color{teal}{\underline{\textbf{8.33\%}}} & 8.73\% & \underline{\textbf{8.34\%}} \\
 \cmidrule{2-17}
 & Param & - & \underline{\color{teal}{\textbf{0.11}}} & 8.07 & 0.30 & 0.37 & 0.31 & \underline{\textbf{0.21}} & 0.40 & 0.75 & \underline{\textbf{0.55}} & 1.35 & 16.42 & 25.47 & 0.27 & {\underline{\textbf{0.12}}} \\
 & Speed & - & \underline{\textbf{11.81}} & 13.01 & \underline{\textbf{14.21}} & 57.67 & 16.88 & 117.09 & 124.82 & 16.04 & 12.01 & \underline{\textbf{7.51}} & 29.38 & 745.62 & 16.87 & \color{teal}{\underline{\textbf{3.54}}} \\
\midrule
\multirow{6}*{\rotatebox{90}{\textbf{PEMS07\ \ }}}  & MAE & 29.64 & \underline{\textbf{22.00}} & 26.14 & 22.25 & 21.16 & 20.44 & 20.04 & \color{teal}{\underline{\textbf{19.49}}} & \underline{\textbf{20.68}} & 21.01 & 22.50 & 22.47 & N/A & 20.59 & \underline{\textbf{19.61}} \\
 & RMSE & 46.23 & \underline{\textbf{35.44}} & 42.72 & 35.83 & 34.15 & 33.38 & 32.86 & \color{teal}{\underline{\textbf{32.59}}} & 34.45 & \underline{\textbf{34.14}} & 36.41 & 35.42 & N/A & 34.86 & \underline{\textbf{32.69}} \\
 & MAPE & 13.52\% & \underline{\textbf{9.31\%}} & 11.37\% & 9.47\% & 9.02\% & 8.71\% & 8.63\% & \color{teal}{\underline{\textbf{8.09\%}}} & \underline{\textbf{8.77\%}} & 8.92\% & 9.57\% & 9.56\% & N/A & 8.61\% & \underline{\textbf{8.31\%}} \\
 & WAPE & 9.47\% & \underline{\textbf{7.03\%}} & 8.36\% & 7.11\% & 6.77\% & 6.54\% & 6.41\% & \color{teal}{\underline{\textbf{6.23\%}}} & \underline{\textbf{6.61\%}} & 6.72\% & 7.19\% & 7.18\% & N/A & 6.58\% & \underline{\textbf{6.27\%}} \\
 \cmidrule{2-17}
 & Param & - & \underline{\color{teal}{\textbf{0.12}}} & 8.07 & 0.59 & 0.37 & 0.32 & \underline{\textbf{0.25}} & 0.41 & \underline{\textbf{0.75}} & 1.37 & 3.44 & 27.18 & N/A & 0.57 & {\underline{\textbf{0.14}}} \\
 & Speed & - & \underline{\textbf{39.30}} & 60.89 & \underline{\textbf{67.14}} & 324.75 & 94.51 & 634.47 & 748.51 & 93.79 & 60.59 & \underline{\textbf{50.93}} & 512.84 & N/A & 73.45 & \color{teal}{\underline{\textbf{13.63}}} \\
\midrule
\multirow{6}*{\rotatebox{90}{\textbf{PEMS08\ \ }}}  & MAE & 21.29 & \underline{\textbf{16.80}} & 18.91 & 16.19 & 15.23 & 14.67 & 14.77 & \underline{\textbf{14.10}} & 15.78 & 15.25 & 16.90 & 16.92 & \color{teal}{\underline{\textbf{14.00}}} & 15.39 & \underline{\textbf{14.21}} \\
 & RMSE & 33.46 & \underline{\textbf{26.38}} & 31.39 & 25.51 & 24.17 & 23.55 & 23.81 & \underline{\textbf{23.36}} & 24.76 & 24.22 & 26.30 & 26.68 & \underline{\textbf{23.41}} & 24.80 & \color{teal}{\underline{\textbf{23.35}}} \\
 & MAPE & 14.34\% & \underline{\textbf{10.66\%}} & 12.11\% & 10.82\% & 10.21\% & 9.46\% & 9.77\% & \underline{\textbf{9.33\%}} & 10.42\% & 10.66\% & 11.89\% & 10.88\% & \underline{\textbf{9.50\%}} & 9.91\% & \color{teal}{\underline{\textbf{9.32\%}}} \\
 & WAPE & 9.15\% & \underline{\textbf{7.24\%}} & 8.14\% & 6.98\% & 6.56\% & 6.32\% & 6.36\% & \underline{\textbf{6.07\%}} & 6.80\% & 6.57\% & 7.28\% & 7.29\% & \color{teal}{\underline{\textbf{6.03\%}}} & 6.63\% & \underline{\textbf{6.12\%}} \\
 \cmidrule{2-17}
 & Param & - & \underline{\color{teal}{\textbf{0.10}}} & 8.07 & 0.23 & 0.37 & 0.31 & \underline{\textbf{0.20}} & 0.39 & \underline{\textbf{0.15}} & 0.35 & 1.15 & 17.25 & 26.56 & 0.20 & {\underline{\textbf{0.12}}} \\
 & Speed & - & 10.62 & \underline{\textbf{8.01}} & \underline{\textbf{7.46}} & 35.36 & 10.34 & 138.65 & 109.42 & 11.33 & 9.57 & \underline{\textbf{6.64}} & 17.49 & 420.86 & 10.98 & \color{teal}{\underline{\textbf{3.38}}} \\
\midrule
\bottomrule
    \end{tabular}
    }
  \end{table*}

\subsection{Delving into Spatial Dependencies}
\label{sec:when_and_how_to_model_spatial_dependencies}

Here, we discuss when and how to model spatial dependencies, and we assess the hypothesis in Section \ref{sec:spatial}.
We select four datasets, where METR-LA~(LA) and PEMS-BAY~(BAY) feature high spatial indistinguishability~(see $r_1$ and $r_2$ in Section \ref{sec:spatial}), while ExchangeRate~(ER) and ETTm1 feature very low spatial indistinguishability.
We choose two baseline models that adopt different approaches to the modeling of the spatial dependencies: STID~\cite{2022STID} and AGCRN~\cite{2020AGCRN}.
STID designs trainable spatial identity embeddings, while AGCRN adopts a GCN-based learning module.
Additionally, we remove the spatial modeling components from each of them, obtaining the variant STID$^*$ without the spatial identities and the variant AGCRN$^*$ with an adjacency matrix set to be the identity matrix.

The results are shown in Table \ref{tab:stid_agcrn}. 
On datasets with significant spatial indistinguishability, the use of both trainable spatial identity embeddings and GCNs can yield significant performance gains.
Conversely, on datasets with low spatial indistinguishability, adding these spatial modeling components degrades performance, suggesting that modeling spatial dependencies (or named cross-dimension dependencies) on these datasets is not necessary.

Based on the above discussion, we conclude that spatial indistinguishability is a strong indicator of spatial dependencies and that we do not always need to model spatial dependencies. 
When there is high spatial indistinguishability in the data, it is purposeful to adopt spatial modeling approaches, \eg GCNs, normalization~\cite{2021STNorm}, and spatial identity~\cite{2022STID}, to improve performance.
In contrast, on datasets with low spatial indistinguishability, designing spatial modeling modules needs to be done with extreme care, as this may cause performance degradation.

\subsection{Performance and Efficiency Benchmarking}
\label{sec:main_results}
So far, we have examined thoroughly the impact of heterogeneity among datasets on the promises of different technical directions and solutions.
{\color{black}We find a strong relationship between model architecture and data characteristics.
Next, we discuss: (i) how to select or design an MTS forecasting solution for a given dataset and (ii) how to choose datasets suitable for evaluating a given MTS forecasting solution, and we (iii) comprehensively analyze the performance and efficiency of existing solutions using rich datasets and (iv) discuss the progress made and noteworthy research directions.

\subsubsection{How to select or design MTS solutions for a given dataset.}

\begin{figure}[t]
  \centering
  \includegraphics[width=0.9\linewidth]{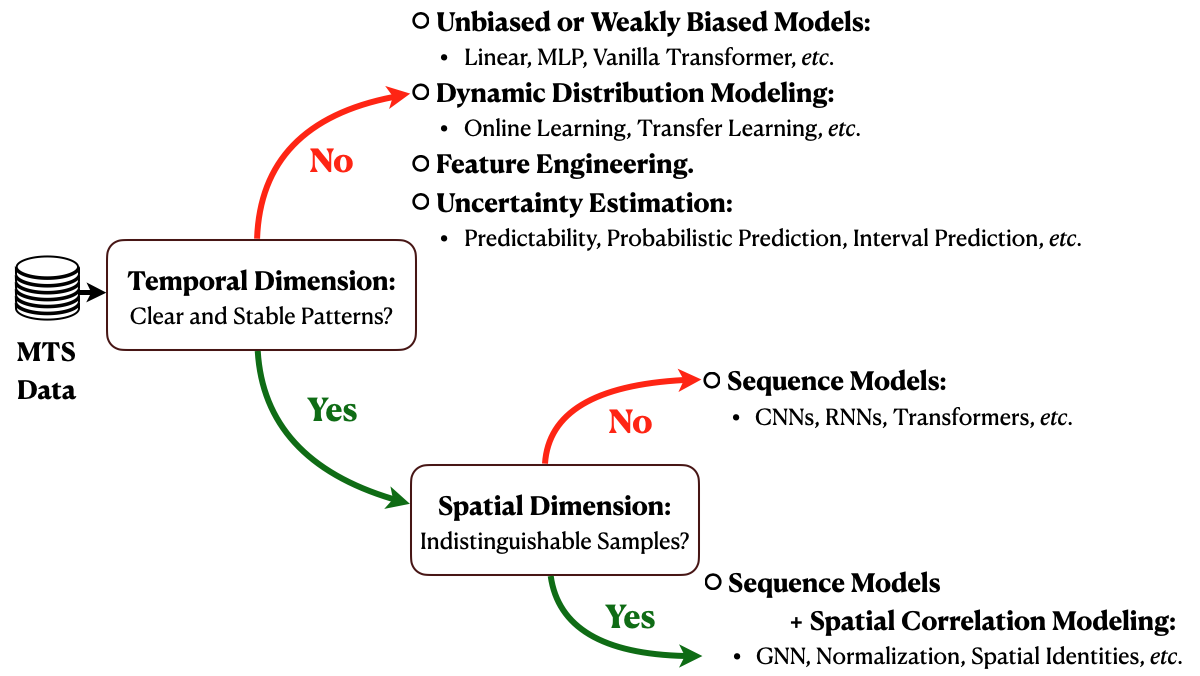}
  \caption{Road map for selecting or designing MTS models.}
  \label{fig:road_map}
\end{figure}

Patterns in the temporal dimension should be examined first. For data exhibiting significant distribution drift or lacking clear patterns, unbiased or weakly biased models should be chosen, \eg linear layers, MLPs, or the vanilla Transformer. If the data displays clear and stable patterns, powerful sequence models are a more reasonable option, \eg TCNs, RNNs, or Transformer architectures. In addition, we need to investigate whether the data has a high sample indistinguishability on the spatial aspect. If so, a spatial dependency modeling module is recommended. Alternative approaches include graph convolution, spatial-temporal normalization, and spatial identity attaching. Moreover, we recommend STID~\cite{2022STID} and Linear~\cite{2023DLinear} as baselines. Given their simplicity, we believe that more complex LTSF or STF solutions are only effective if they can significantly outperform these two. We summarize the above discussion in the road map in Figure \ref{fig:road_map}.

\begin{table*}[htpb]
\setlength\tabcolsep{3.7pt}
\renewcommand\arraystretch{0.5}
    \centering
    \caption{LTSF on PEMS04, PEMS08, ETTh1, ETTm1, Electricity, Weather, and ExchangeRate~(ER) datasets. }
    \label{tab:main_ltsf}
    \scalebox{0.73}{
    \begin{tabular}{cc|ccc|ccccccc|ccc}
\toprule
\midrule
      \textbf{Data} & \textbf{Metrics} & \textbf{LGBM} & \textbf{DeepAR} & \textbf{NBeats} & \textbf{Informer} & \textbf{Autoformer} & \textbf{Pyraformer} & \textbf{FEDformer}& \textbf{Triformer} & \textbf{Crossformer} & \textbf{PatchTST} & \textbf{Linear} & \textbf{DLinear} & \textbf{NLinear}\\
      \midrule
\multirow{5}*{\rotatebox{90}{\textbf{PEMS04\ }}}  & MAE & 34.55 & 34.79 & \underline{\textbf{27.95}} & 27.94 & 34.72 & 25.49 & 26.89 & \color{teal}{\underline{\textbf{23.81}}} & 26.75 & 25.72 & \underline{\textbf{37.42}} & 37.52 & 37.62 \\
 & RMSE & 57.74 & 55.91 & \underline{\textbf{46.87}} & 44.74 & 50.33 & 41.74 & 41.46 & \color{teal}{\underline{\textbf{39.42}}} & 45.24 & 40.13 & \underline{\textbf{62.14}} & 62.21 & 62.38 \\
 & WAPE & 14.94\% & 15.83\% & \underline{\textbf{12.86\%}} & 12.84\% & 14.81\% & 11.72\% & 12.39\% & \color{teal}{\underline{\textbf{10.95\%}}} & 12.31\% & 13.35\% & \underline{\textbf{17.22\%}} & 17.26\% & 11.35\% \\
 \cmidrule{2-15}
 & Param & - & \color{teal}{\underline{\textbf{0.11}}} & 13.05 & 12.40 & 12.07 & 218.42 & 18.36 & \underline{\textbf{1.69}} & 13.51 & 2.34 & \color{teal}{\underline{\textbf{0.11}}} & 0.23 & \color{teal}{\underline{\textbf{0.11}}} \\
 & Speed & - & 392.98 & \underline{\textbf{31.65}} & 28.62 & 137.61 & \underline{\textbf{25.22}} & 147.75 & 313.52 & 338.94 & 147.42 & \color{teal}{\underline{\textbf{21.13}}} & 21.24 & 21.27 \\
\midrule
\multirow{5}*{\rotatebox{90}{\textbf{PEMS08\ }}}  & MAE & 38.15 & 35.58 & \underline{\textbf{21.43}} & 26.92 & 33.75 & 22.03 & 25.14 & \color{teal}{\underline{\textbf{18.74}}} & 21.75 & 19.86 & \underline{\textbf{34.04}} & 34.15 & 34.11 \\
 & RMSE & 57.74 & 54.98 & \underline{\textbf{38.69}} & 43.79 & 51.09 & 38.39 & 39.17 & \color{teal}{\underline{\textbf{31.03}}} & 36.86 & 33.44 & \underline{\textbf{57.07}} & 57.18 & 57.26 \\
 & WAPE & 14.94\% & 13.14\% & \underline{\textbf{9.26\%}} & 11.63\% & 15.37\% & 9.52\% & 10.87\% & \color{teal}{\underline{\textbf{8.13\%}}} & 9.40\% & 8.34\% & \underline{\textbf{14.71\%}} & 14.76\% & 14.74\% \\
 \cmidrule{2-15}
 & Param & - & \underline{\color{teal}\textbf{0.10}} & 13.05 & 11.91 & 11.37 & 123.93 & 17.66 & \underline{\textbf{1.64}} & 12.53 & 2.34 & {\underline{\textbf{0.11}}} & 0.23 & {\underline{\textbf{0.11}}} \\
 & Speed & - & 557.72 & \underline{\textbf{18.06}} & 23.34 & 81.54 & \underline{\textbf{20.94}} & 147.89 & 200.50 & 224.90 & 86.81 & \color{teal}{\underline{\textbf{12.60}}} & 12.78 & 13.04 \\
\midrule
\multirow{5}*{\rotatebox{90}{\textbf{ETTh1\ }}}  & MAE & \underline{\textbf{1.76}} & 1.94 & 1.83 & 2.92 & 1.74 & 2.68 & 1.71 & 1.80 & 1.83 & \underline{\textbf{1.60}} & 1.60 & \color{teal}{\underline{\textbf{1.58}}} & 1.59 \\
 & RMSE & \underline{\textbf{3.34}} & 3.44 & 3.36 & 4.60 & 3.12 & 4.26 & 3.15 & 3.31 & 3.19 & \underline{\textbf{3.08}} & 3.08 & \color{teal}{\underline{\textbf{3.06}}} & 3.09 \\
 & WAPE & \underline{\textbf{38.30\%}} & 41.89\% & 39.50\% & 62.87\% & 37.61\% & 57.75\% & 36.89\% & 38.72\% & 39.44\% & \underline{\textbf{34.49\%}} & 34.47\% & \color{teal}{\underline{\textbf{34.06\%}}} & 34.29\% \\
 \cmidrule{2-15}
 & Param & - & \underline{\color{teal}\textbf{0.10}} & 13.05 & 11.33 & 10.54 & 11.59 & 16.30 & 1.59 & 11.36 & \underline{\textbf{0.24}} & {\underline{\textbf{0.11}}} & 0.23 & {\underline{\textbf{0.11}}} \\
 & Speed & - & 110.53 & \underline{\textbf{3.58}} & 14.18 & 39.37 & 11.98 & 43.44 & 26.24 & 11.91 & \underline{\textbf{3.90}} & \color{teal}{\underline{\textbf{1.25}}} & 1.35 & 1.28 \\
\midrule
\multirow{5}*{\rotatebox{90}{\textbf{ETTm1\ }}}  & MAE & 1.54 & 2.21 & \underline{\textbf{1.53}} & 2.37 & 1.93 & 2.45 & 1.53 & 1.57 & 1.73 & \color{teal}{\underline{\textbf{1.37}}} & 1.39 & \underline{\textbf{1.38}} & \underline{\textbf{1.38}} \\
 & RMSE & \underline{\textbf{2.92}} & 4.02 & 2.94 & 4.20 & 3.67 & 3.92 & 2.89 & 2.94 & 3.01 & \color{teal}{\underline{\textbf{2.78}}} & \underline{\textbf{2.80}} & \underline{\textbf{2.80}} & 2.81 \\
 & WAPE & 33.36\% & 47.72\% & \underline{\textbf{33.03\%}} & 51.16\% & 41.69\% & 52.79\% & 33.00\% & 33.83\% & 37.52\% & \color{teal}{\underline{\textbf{29.60\%}}} & 30.02\% & \underline{\textbf{29.80\%}} & 29.86\% \\
 \cmidrule{2-15}
 & Param & - & \underline{\color{teal}\textbf{0.10}} & 13.05 & 11.33 & 10.54 & 11.59 & 16.83 & 1.59 & 11.79 & \underline{\textbf{0.24}} & {\underline{\textbf{0.11}}} & 0.23 & {\underline{\textbf{0.11}}} \\
 & Speed & - & 440.85 & \underline{\textbf{8.57}} & 57.32 & 121.91 & 46.57 & 235.17 & 108.11 & 98.13 & \underline{\textbf{14.77}} & \color{teal}{\underline{\textbf{4.35}}} & 4.79 & 4.40 \\
\midrule
\multirow{5}*{\rotatebox{90}{\textbf{Electricity}}}  & MAE & 543.11 & 283.72 & \underline{\textbf{281.63}} & 325.53 & 295.98 & 335.23 & 317.20 & 275.42 & 283.86 & \underline{\textbf{253.57}} & 256.60 & \color{teal}{\underline{\textbf{250.08}}} & 251.80 \\
 & RMSE & 8352.53 & 2998.18 & \underline{\textbf{2936.83}} & 2938.56 & 2933.97 & \color{teal}{\underline{\textbf{2761.75}}} & 2935.53 & 2968.97 & 2732.70 & 2881.53 & 2896.04 & \underline{\textbf{2883.63}} & 2892.13 \\
 & WAPE & 20.37\% & 12.27\% & \underline{\textbf{10.56\%}} & 12.22\% & 11.11\% & 12.58\% & 11.91\% & 10.33\% & 10.56\% & \underline{\textbf{9.51\%}} & 9.62\% & \color{teal}{\underline{\textbf{9.38\%}}} & 9.44\% \\
 \cmidrule{2-15}
 & Param & - & \underline{\color{teal}\textbf{0.11}} & 13.05 & 12.45 & 12.14 & 228.20 & 17.91 & 1.69 & 1.20 & \underline{\textbf{0.24}} & \color{teal}{\underline{\textbf{0.11}}} & 0.23 & \color{teal}{\underline{\textbf{0.11}}} \\
 & Speed & - & 918.78 & \underline{\underline{\textbf{55.53}}} & \underline{\textbf{51.00}} & 69.04 & 70.86 & 113.94 & 484.90 & 198.54 & 95.92 & 40.92 & \color{teal}{\underline{\textbf{39.21}}} & 40.71 \\
\midrule
\multirow{5}*{\rotatebox{90}{\textbf{Weather\ }}}  & MAE & 16.66 & 24.88 & \underline{\textbf{12.29}} & 12.88 & 29.37 & 38.94 & 15.61 & 11.29 & 11.36 & \color{teal}{\underline{\textbf{10.85}}} & 12.25 & 12.08 & \underline{\textbf{12.02}} \\
 & RMSE & 73.78 & 73.44 & \underline{\textbf{44.57}} & 41.57 & 84.75 & 142.50 & 44.74 & \color{teal}{\underline{\textbf{40.74}}} & 41.27 & 41.70 & \underline{\textbf{43.22}} & 43.35 & 43.32 \\
 & WAPE & 9.96\% & 14.87\% & \underline{\textbf{7.29\%}} & 7.70\% & 17.56\% & 23.28\% & 9.33\% & 6.75\% & 6.79\% & \color{teal}{\underline{\textbf{6.49\%}}} & 7.32\% & 7.22\% & \underline{\textbf{7.19\%}} \\
 \cmidrule{2-15}
 & Param & - & \underline{\color{teal}\textbf{0.10}} & 13.05 & 11.38 & 10.61 & 21.30 & 16.90 & \underline{\textbf{1.59}} & 11.46 & 2.21 & {\underline{\textbf{0.11}}} & 0.23 & {\underline{\textbf{0.11}}} \\
 & Speed & - & 439.89 & \underline{\textbf{10.77}} & 78.30 & 99.51 & 50.52 & 503.17 & 120.21 & 107.23 & \underline{\textbf{42.72}} & \color{teal}{\underline{\textbf{8.19}}} & 8.73 & 8.38 \\
\midrule
\multirow{5}*{\rotatebox{90}{\textbf{ER\quad \ }}} & MAE & 0.0940 & 0.0608 & \underline{\textbf{0.0342}} & 0.0611 & 0.0366 & 0.0632 & 0.0376 & 0.0367 & 0.0504 & \underline{\textbf{0.0332}} & 0.0352 & 0.0350 & \color{teal}{\underline{\textbf{0.0322}}} \\
 & RMSE & 0.1531 & 0.0885 & \underline{\textbf{0.0537}} & 0.0803 & 0.0568 & 0.0870 & 0.0578 & 0.0533 & 0.0732 & \underline{\textbf{0.0525}} & 0.0550 & 0.0547 & \color{teal}{\underline{\textbf{0.0508}}} \\
& WAPE & 11.55\% & 8.00\% & \underline{\textbf{4.51\%}} & 8.0355\% & 4.8152\% & 8.3192\% & 4.9405\% & 4.8431\% & 6.6406\% & \underline{\textbf{4.3804\%}} & 4.6343\% & 4.6086\% & \color{teal}{\underline{\textbf{4.2450\%}}} \\
\cmidrule{2-15}
& Param & - & \underline{\color{teal}\textbf{0.10}} & 13.05 & 11.33 & 10.54 & 12.26 & 16.31 & 1.59 & 0.78 & \underline{\textbf{0.24}} & {\underline{\textbf{0.11}}} & 0.23 & {\underline{\textbf{0.11}}} \\
& Speed & - & 64.02 & \underline{\textbf{1.39}} & 8.60 & 13.72 & 13.03 & 26.20 & 15.98 & 6.65 & \underline{\textbf{2.38}} & 0.93 & 0.95 & \color{teal}{\underline{\textbf{0.91}}} \\
\midrule
\bottomrule
\end{tabular}
}
\end{table*}

\subsubsection{How to choose suitable datasets for evaluating a given MTS solution.}
The key to validating the effectiveness of solutions, which are usually designed to address specific tasks, is to select datasets that align with the task objectives. For instance, STF algorithms often aim to model spatial-temporal dependencies. Thus, datasets with significant spatial dependency are necessary to validate the spatial modeling. LTSF algorithms, on the other hand, aim for generic time series forecasting and should be validated on datasets with and without clear and stable patterns to assess their generalization. However, most LTSF studies only validate on datasets lacking clear patterns like ETT or ExchangeRate. Our experimental results show that this can create an illusion of progress.

Moreover, there are times when our objective is practical, ranking the performance of popular algorithms. In such cases, real compound data is more suitable as it typically encompasses multiple challenges simultaneously. For example, the M4 competition dataset comprises both time series with and without clear and stable patterns. It is important to note that a solution designed specifically for one type of task might not outperform others on such datasets as it contains multiple complex tasks. For instance, SOTA models in STF or LTSF might not yield satisfactory results on the M4 dataset.
}

\subsubsection{Experimental Results}

First, we present and discuss the detailed performance and efficiency evaluations on LTSF and STF tasks. 
Then, we select representative solutions from STF and LTSF, along with classic time series solutions, and showcase their results on the complex competition M4 dataset.

The results for LTSF are shown in Table \ref{tab:main_ltsf}.
 {\color{black}When used on datasets without clear and stable patterns, the state-of-the-art advanced Transformer models~\cite{2023Crossformer,2023PatchTST} and the basic linear models~\cite{2023DLinear} exhibit comparable performance. 
Considering the simplicity of Linear-based models, \textit{we believe that for LTSF prediction tasks, designing new training strategies or engaging in feature engineering to address distribution drift or ambiguous patterns poses more important challenges than designing increasingly more complex time series forecasting models.}}
Moreover, on datasets with clear and stable patterns, it is surprising that many recent solutions struggle to outperform the earliest baseline, Informer~\cite{2021Informer}. 
Considering that making predictions on such datasets should be a more straightforward task, this raises concerns that the architectures of existing LTSF models might have been over-fitted datasets like ETT, Electricity, Weather, and ExchangeRate that are used commonly in LTSF studies. 
This reaffirms the importance of selecting appropriate evaluation datasets.

Table \ref{tab:main_stf} reports the experimental results for STF. 
Benefiting from the incorporation of prior knowledge, prior-graph-based methods generally perform better than latent-graph-based or non-graph-based methods.
Furthermore, it is apparent that learning a graph structure can be very challenging.
Among the different solutions, only MTGNN~\cite{2020MTGNN} and STEP~\cite{2022STEP} are capable of learning effective graph structures that do not significantly degrade the prediction performance.
\textit{Overall, it is obvious that more intricate network structures yield very limited improvement. }
For example, although D$^2$STGNN~\cite{2022D2STGNN}, published in 2022, is the state-of-the-art for STF prediction, its MAE on METR-LA is only 6\% higher than that of Graph WaveNet~\cite{2019GWNet}, published in 2019. 
In addition, it is even more surprising that Graph WaveNet~\cite{2019GWNet} and its variant MTGNN are still able to significantly outperform many newer solutions, including StemGNN~\cite{2020StemGNN}, GTS~\cite{2021GTS}, and others~\cite{STGODE, 2021Z-GCNets, 2022STG-NCDE}.
Therefore, \textit{we find that compared to improving prediction accuracy by designing increasingly complex models, more progress may be achieved by focusing on other important and challenging issues, such as efficiency, graph structure learning.} For example, STID and STNorm are highly efficient and have achieved satisfactory results on most datasets.

In summary, advanced solutions for LTSF and STF represent substantial progress on the modeling of long-term time dependencies and spatial dependencies, respectively.
However, complex industrial datasets often contain more complex challenges.
Table \ref{tab:m4} reports the experimental results of representative solutions on the M4 dataset.
Specifically, LGBM, DeepAR, and NBeats are widely used solutions in industrial applications; STID represents STF prediction solutions, while PatchTST represents LTSF solutions. We follow an existing experimental setup from~\cite{2022TimesNet} and report their results on the Yearly, Quarterly, Monthly, and Others subsets, including also their weighted averages.
As in the literature~\cite{2022TimesNet}, we remove the ensemble method in NBeats for fair comparison.
Although PatchTST and STID are superior in Tables \ref{tab:main_ltsf} and \ref{tab:main_stf}, they perform worse than classic algorithms on the M4 dataset.

\begin{table}[t]
\setlength\tabcolsep{7.5pt}
\renewcommand\arraystretch{0.8}
\caption{Results on the M4 dataset.}
\centering
\label{tab:m4}
\scalebox{0.75}{
\begin{tabular}{cc|ccc|cc}
\toprule
\midrule
\multicolumn{2}{c}{\textbf{Models}} & \textbf{LGBM} & \textbf{DeepAR} & \multicolumn{1}{c}{\textbf{NBeats}} & \multicolumn{1}{c}{\textbf{STID}} & \textbf{PatchTST} \\
\midrule
\midrule
\multicolumn{1}{c}{\multirow{3}{*}{\rotatebox{90}{\scriptsize\textbf{Yearly}}}} & \multicolumn{1}{|c|}{sMAPE} &14.705 & 13.886 & \color{teal}\underline{\textbf{13.337}} & \underline{\textbf{13.420}} & 14.158\\
& \multicolumn{1}{|c|}{MASE} & 3.565 & 3.129 & \color{teal}\underline{\textbf{3.004}} & \underline{\textbf{3.071}} & 3.193 \\
& \multicolumn{1}{|c|}{OWA} & 0.898 & 0.819 & \color{teal}\underline{\textbf{0.786}} & \underline{\textbf{0.797}} & 0.836\\
\midrule
\multicolumn{1}{c}{\multirow{3}{*}{\rotatebox{90}{\textbf{\scriptsize{Quarterly}}}}} & \multicolumn{1}{|c|}{sMAPE} & 11.358 & 11.374 & \color{teal}\underline{\textbf{9.866}} & \underline{\textbf{9.869}} & 10.257\\
& \multicolumn{1}{|c|}{MASE} & 1.418 & 1.352 & \color{teal}\underline{\textbf{1.136}} & \underline{\textbf{1.141}} & 1.184\\
& \multicolumn{1}{|c|}{OWA} & 1.033 & 1.009 & \color{teal}\underline{\textbf{0.862}} & \underline{\textbf{0.864}} & 0.898\\
\midrule
\multicolumn{1}{c}{\multirow{3}{*}{\rotatebox{90}{\textbf{\scriptsize{Monthly}}}}} & \multicolumn{1}{|c|}{sMAPE} & 14.559 & 14.749 & \color{teal}\underline{\textbf{12.168}} & \underline{\textbf{12.624}} & 13.244\\
& \multicolumn{1}{|c|}{MASE} & 1.172 & 1.185 & \color{teal}\underline{\textbf{0.897}} & \underline{\textbf{0.940}} & 1.002\\
& \multicolumn{1}{|c|}{OWA} & 1.056 & 1.068 & \color{teal}\underline{\textbf{0.844}} & \underline{\textbf{0.879}} & 0.93\\
\midrule
\multicolumn{1}{c}{\multirow{3}{*}{\rotatebox{90}{\scriptsize\textbf{Others}}}} & \multicolumn{1}{|c|}{sMAPE} & 6.665 & 6.410 & \color{teal}\underline{\textbf{4.635}} & \underline{\textbf{4.806}} & 4.844\\
& \multicolumn{1}{|c|}{MASE} & 4.810 & 4.769 & \color{teal}\underline{\textbf{3.106}} & 3.358 & \underline{\textbf{3.166}}\\
& \multicolumn{1}{|c|}{OWA} & 1.460 & 1.427 & \color{teal}\underline{\textbf{0.978}} & 1.035 & \underline{\textbf{1.009}}\\
\midrule
\multicolumn{1}{c}{\multirow{3}{*}{\rotatebox{90}{\makecell{\textbf{\scriptsize{Weighted}}\\\textbf{\scriptsize{Average}}}}}} & \multicolumn{1}{|c|}{sMAPE} & 13.430 & 13.323 & \color{teal}\underline{\textbf{11.508}} & \underline{\textbf{11.755}} & 12.324\\
& \multicolumn{1}{|c|}{MASE} & 1.963 & 1.851 & \color{teal}\underline{\textbf{1.550}} & \underline{\textbf{1.599}} & 1.658\\
& \multicolumn{1}{|c|}{OWA} & 1.008 & 0.975 & \color{teal}\underline{\textbf{0.829}} & \underline{\textbf{0.851}} & 0.888\\
\midrule
\bottomrule

\end{tabular}
}
\end{table}

\subsubsection{Limitations of Current Studies and Future Directions}{\color{black}There is no doubt that multivariate time series hold significant value in various scientific fields~\cite{InnovationGeo, AI4Science, MGSFformer}. Although deep learning-based MTS forecasting solutions, particularly in STF and LTSF, have seen substantial advancements, current efforts mainly focus on designing increasingly intricate model architectures. The limitation is that these endeavors appear to be effective only when the data exhibits strong patterns. 
However, unlike image~\cite{Image1, Image2} and natural language data, whose patterns are frequently consistent and stable, time series data can be greatly affected by external factors, resulting in distribution drift or the frequent occurrence of unpredictable changes. Moreover, MTS data in real-world scenarios often face challenges related to insufficient data volume and low data quality~\cite{GinAR}. These factors represent key bottlenecks for the broader application of existing research outcomes. Therefore, we emphasize that future research should prioritize more realistic scenarios, such as modeling distribution shifts, predicting with low-quality data, and zero- or few-shot learning.
}

\section{Conclusion}
\label{sec:conclusions}
In this study, we address the seemingly inconsistent experimental findings and difficulties in selecting technical directions in the area of Multivariate Time Series~(MTS) forecasting, shedding light on the actual advance achieved.
First, we introduce a novel benchmark called \BasicTS that is designed to enable fair and reasonable comparisons of MTS forecasting solutions.
By adopting a unified training pipeline, \BasicTS addresses the issue of inconsistent performance, and provides more reasonable evaluation procedures.
Second, we delve into the heterogeneity across MTS datasets.
Considering the temporal aspect, we categorize datasets according to whether they exhibit clear and stable patterns, significant distribution drift, or unclear patterns.
Considering the spatial aspect, we devise metrics to quantify spatial dependencies and partition datasets into those with and without significant spatial indistinguishability.
We emphasize that many conclusions drawn in prior research hold only for certain types of data, and considering these conclusions to be more general can lead researchers to make counterproductive inferences.
Additionally, using \BasicTS and the associated MTS datasets, we conduct an extensive analysis and comparison of popular solutions.
These findings offer valuable insight into the progress already made, aiding researchers in choosing appropriate solutions or datasets and drawing more reliable conclusions.

\section*{Acknowledgments}
{\color{black}
This work is supported by the NSFC under Grant Nos. 62372430, 62206266, 62476264, and 62472405, and by the Youth Innovation Promotion Association of CAS under Grant No. 2023112. This work is also funded by the Postdoctoral Fellowship Program of CPSF under Grant No. GZC20241758.
}


{\appendix

\subsection{Datasets}

\begin{table}[b]
\setlength\tabcolsep{4pt}
\setstretch{1.0}
\caption{Statistics of datasets.}
\label{tab:datasets}
\scalebox{0.87}{
  \begin{tabular}{|c|c|c|c|c|c|}
    \hline
    \textbf{Dataset} &\textbf{Samples} & \textbf{Variates} & \textbf{Frequency} & \textbf{Time Span} & \textbf{Graph}\\
    \hline
    \textbf{METR-LA}  & 34272 & 207 & 5 mins & 4 months & Yes\\
    \hline
    \textbf{PEMS-BAY} & 52116 & 325 & 5 mins & 6 months & Yes\\
    \hline
    \textbf{PEMS03}   & 26208 & 358 & 5 mins & 3 months & Yes\\
    \hline
    \textbf{PEMS04}   & 16992 & 307 & 5 mins & 2 months & Yes\\
    \hline
    \textbf{PEMS07}   & 28224 & 883 & 5 mins & 4 months & Yes\\
    \hline
    \textbf{PEMS08}   & 17856 & 170 & 5 mins & 2 months & Yes\\
    \hline
    \textbf{ETTh1}   & 14400 & 7 & 1 hour & 20 months & No\\
    \hline
    \textbf{ETTh2}   & 14400 & 7 & 1 hour & 20 months & No\\
    \hline
    \textbf{ETTm1}   & 57600 & 7 & 15 mins & 20 months & No\\
    \hline
    \textbf{ETTm2}   & 57600 & 7 & 15 mins & 20 months & No\\
    \hline
    \textbf{Electricity}   & 26304 & 321 & 1 hour & 2 years & No\\
    \hline
    \textbf{Weather}   & 52696 & 21 & 10 mins & 1 year & No\\
    \hline
    \textbf{\makecell{Exchange\\Rate}} & 7588 & 8 & 1 day & 27 Years & No\\
    \hline
    {\color{black}\textbf{M4}} & {\color{black}19-9933} & {\color{black}100000} & {\color{black}Mixed} & {\color{black}N/A} & {\color{black}No}\\
    \hline
  \end{tabular}
  }
\end{table}

\begin{itemize}
\item \textbf{METR-LA} and \textbf{PEMS-BAY} are traffic speed datasets recorded every {5} minutes. The datasets include sensor graphs to indicate spatial dependencies between sensors~\cite{2018DCRNN}.

\item \textbf{PEMS03, PEMS04, PEMS07, and PEMS08} are traffic flow datasets recorded every {5} minutes. These datasets include sensor graphs to indicate dependencies between sensors~\cite{2019ASTGCN}.

\item \textbf{ETTh1, ETTh2, ETTm1, and ETTm2} record temperatures of electricity transformers  for use in electric power long-term deployments. The \textquoteleft 1\textquoteright and \textquoteleft 2\textquoteright indicate different transformers, while \textquoteleft h \textquoteright and \textquoteleft m\textquoteright indicate different sampling every hour and every 15 minutes~\cite{2021Informer}.

\item \textbf{Electricity} records electricity consumption in kWh by 321 clients every hour from 2012 to 2014~\cite{2018LSTNet}.

\item \textbf{Weather} records 21 meteorological indicators every 10 minutes for the year of 2020~\cite{2021AutoFormer}. 

\item \textbf{ExchangeRate} collects the daily exchange rates for the currencies of eight countries including Australia, UK, Canada, Switzerland, China, Japan, New Zealand, and Singapore~\cite{2018LSTNet}.

\item \textbf{M4} is a well-known competition dataset that contains marketing data collected yearly, quarterly, monthly, weekly, daily, and hourly, with a total of 100,000 time series. These time series are not aligned, which means they have different lengths and start and end times~\cite{M4team2018dataset}.
\end{itemize}
Table \ref{tab:datasets} reports summary statistics on the datasets.

\subsection{Baselines}

Table \ref{tab:baselines} gives a brief summary of the baselines used in this study.

\begin{table}[ht]
\setlength\tabcolsep{3pt}
\setstretch{1.0}
\caption{Overview of baselines.}
\label{tab:baselines}
\scalebox{0.87}{
  \begin{tabular}{|c|c|c|c|}
    \hline
    \textbf{Methods} & \textbf{Fied} & \textbf{Category} & \textbf{Notes}\\
    \hline
    \textbf{STGCN} & \multirow{12}*{{STF}} & \multirow{5}*{{\makecell{Prior-\\Graph-\\based}}} & GCN + TCN\\
    \cline{1-1}\cline{4-4}
    \textbf{DCRNN}  &  &  & GCN + RNN\\
    \cline{1-1}\cline{4-4}
    \textbf{GWNet}  & &  & GCN + TCN\\
     \cline{1-1}\cline{4-4}
    \textbf{DGCRN}  &   &  &GCN + RNN\\
     \cline{1-1}\cline{4-4}
    \textbf{D$^2$STGNN}   &  &  & Mixed\\
    \cline{1-1} \cline{3-4}
    \textbf{AGCRN}  &   & \multirow{5}*{{\makecell{Latent-\\Graph-\\based}}} & GCN + RNN\\
    \cline{1-1}\cline{4-4}
    \textbf{MTGNN}   &  &  & GCN + TCN\\
    \cline{1-1}\cline{4-4}
    \textbf{StemGNN}  &   &  & GCN + TCN + Spectral\\
    \cline{1-1}\cline{4-4}
    \textbf{GTS}  &   &  & GCN + RNN\\
    \cline{1-1}\cline{4-4}
    \textbf{STEP}  &   &  & Pretraining-Enhanced\\
    \cline{1-1} \cline{3-4}
    \textbf{ST-Norm}   &  & \multirow{2}*{{\makecell{\small Non-Graph-\\based}}} & Normalization\\
    \cline{1-1}\cline{4-4}
    \textbf{STID}   &  &  & Identity Embedding\\
    \hline
    \textbf{Informer}  & \multirow{9}*{{LTSF}} & \color{black}\multirow{6}*{{\makecell{\small Advanced\\\small Neural\\\small Networks}}} & Efficient Self-Attention\\
    \cline{1-1}\cline{4-4}
    \textbf{Autoformer}  &  &  & Auto-Correlation\\
    \cline{1-1}\cline{4-4}
    \textbf{FEDformer}  &  &  & Frequency-Enhanced\\
    \cline{1-1}\cline{4-4}
    \textbf{Pyraformer}  &  &  & Pyramidal Attention\\
    \cline{1-1}\cline{4-4}
    \textbf{Triformer} & & & Triangular Attention \\
    \cline{1-1}\cline{4-4}
    \textbf{Crossformer}  &  &  & Cross-Dimension Attention\\
    \cline{1-1}\cline{4-4}
    \textbf{PatchTST}  &  &  & Channel Independent + Patchify\\
    \cline{1-1} \cline{3-4}
    \textbf{Linear}  &  & \color{black}\multirow{3}*{{\makecell{
    \vspace{-0.1cm}\small Basic\\ \vspace{-0.1cm} \small Neural\\ \vspace{-0.1cm} \small Networks
    }}} & Vanilla Linear Layer\\
    \cline{1-1}\cline{4-4}
    \textbf{NLinear}  &  &  & Linear + Normalization\\
    \cline{1-1}\cline{4-4}
    \textbf{DLinear}  &  &  & Linear + Decomposition\\
    \hline
    \textbf{LGBM} &\multicolumn{2}{c|}{\multirow{3}{*}{Classic}} & Gradient Boosting\\
    \cline{1-1}\cline{4-4}
    \textbf{DeepAR} &\multicolumn{2}{c|}{} & Probabilistic Auto Regressive\\
    \cline{1-1}\cline{4-4}
    \textbf{NBeats} &\multicolumn{2}{c|}{} & Neural Basis Expansion\\
    \hline
  \end{tabular}
  }
\end{table}

}

\bibliographystyle{IEEEtrans}
\normalem
\bibliography{references_nopp}

\begin{thebibliography}{10}
\providecommand{\url}[1]{#1}
\csname url@samestyle\endcsname
\providecommand{\newblock}{\relax}
\providecommand{\bibinfo}[2]{#2}
\providecommand{\BIBentrySTDinterwordspacing}{\spaceskip=0pt\relax}
\providecommand{\BIBentryALTinterwordstretchfactor}{4}
\providecommand{\BIBentryALTinterwordspacing}{\spaceskip=\fontdimen2\font plus
\BIBentryALTinterwordstretchfactor\fontdimen3\font minus \fontdimen4\font\relax}
\providecommand{\BIBforeignlanguage}[2]{{%
\expandafter\ifx\csname l@#1\endcsname\relax
\typeout{** WARNING: IEEEtran.bst: No hyphenation pattern has been}%
\typeout{** loaded for the language `#1'. Using the pattern for}%
\typeout{** the default language instead.}%
\else
\language=\csname l@#1\endcsname
\fi
#2}}
\providecommand{\BIBdecl}{\relax}
\BIBdecl

\bibitem{PEMS-BAY}
C.~Chen, K.~Petty, A.~Skabardonis, P.~Varaiya, and Z.~Jia, ``Freeway performance measurement system: mining loop detector data,'' \emph{Transportation Research Record}, vol. 1748, no.~1, 2001.

\bibitem{2018LSTNet}
G.~Lai, W.~Chang, Y.~Yang, and H.~Liu, ``Modeling long- and short-term temporal patterns with deep neural networks,'' in \emph{SIGIR}.\hskip 1em plus 0.5em minus 0.4em\relax {ACM}, 2018.

\bibitem{2022D2STGNN}
Z.~Shao, Z.~Zhang, W.~Wei, F.~Wang, Y.~Xu, X.~Cao, and C.~S. Jensen, ``Decoupled dynamic spatial-temporal graph neural network for traffic forecasting,'' \emph{Proc. {VLDB} Endow.}, vol.~15, no.~11, 2022.

\bibitem{Nature}
H.~Wu, H.~Zhou, M.~Long, and J.~Wang, ``Interpretable weather forecasting for worldwide stations with a unified deep model,'' \emph{Nature Machine Intelligence}, 2023.

\bibitem{2023SUFS}
L.~Sun, S.~Gong, T.~Zhang, F.~Jiang, Z.~Zhao, J.~Chen, and X.~Zhang, ``{SUFS:} {A} generic storage usage forecasting service through adaptive ensemble learning,'' in \emph{ICDE}.\hskip 1em plus 0.5em minus 0.4em\relax {IEEE}, 2023.

\bibitem{2018DCRNN}
Y.~Li, R.~Yu, C.~Shahabi, and Y.~Liu, ``Diffusion convolutional recurrent neural network: Data-driven traffic forecasting,'' in \emph{ICLR}, 2018.

\bibitem{2021Informer}
H.~Zhou, S.~Zhang, J.~Peng, S.~Zhang, J.~Li, H.~Xiong, and W.~Zhang, ``Informer: Beyond efficient transformer for long sequence time-series forecasting,'' in \emph{AAAI}, 2021.

\bibitem{2023Crossformer}
Y.~Zhang and J.~Yan, ``Crossformer: Transformer utilizing cross-dimension dependency for multivariate time series forecasting,'' in \emph{ICLR}, 2023.

\bibitem{2021AutoFormer}
H.~Wu, J.~Xu, J.~Wang, and M.~Long, ``Autoformer: Decomposition transformers with auto-correlation for long-term series forecasting,'' in \emph{NeurIPS}, 2021.

\bibitem{2022FEDFormer}
T.~Zhou, Z.~Ma, Q.~Wen, X.~Wang, L.~Sun, and R.~Jin, ``Fedformer: Frequency enhanced decomposed transformer for long-term series forecasting,'' in \emph{ICML}, vol. 162.\hskip 1em plus 0.5em minus 0.4em\relax {PMLR}, 2022.

\bibitem{2019GWNet}
Z.~Wu, S.~Pan, G.~Long, J.~Jiang, and C.~Zhang, ``Graph wavenet for deep spatial-temporal graph modeling,'' in \emph{IJCAI}.\hskip 1em plus 0.5em minus 0.4em\relax ijcai.org, 2019.

\bibitem{2018STGCN}
B.~Yu, H.~Yin, and Z.~Zhu, ``Spatio-temporal graph convolutional networks: {A} deep learning framework for traffic forecasting,'' in \emph{IJCAI}, 2018.

\bibitem{2017Transformer}
A.~Vaswani, N.~Shazeer, N.~Parmar, J.~Uszkoreit, L.~Jones, A.~N. Gomez, L.~Kaiser, and I.~Polosukhin, ``Attention is all you need,'' in \emph{NeurIPS}, 2017.

\bibitem{2019LogTrans}
S.~Li, X.~Jin, Y.~Xuan, X.~Zhou, W.~Chen, Y.~Wang, and X.~Yan, ``Enhancing the locality and breaking the memory bottleneck of transformer on time series forecasting,'' in \emph{NeurIPS}, 2019.

\bibitem{2022Pyraformer}
S.~Liu, H.~Yu, C.~Liao, J.~Li, W.~Lin, A.~X. Liu, and S.~Dustdar, ``Pyraformer: Low-complexity pyramidal attention for long-range time series modeling and forecasting,'' in \emph{ICLR}, 2022.

\bibitem{2023PatchTST}
Y.~Nie, N.~H. Nguyen, P.~Sinthong, and J.~Kalagnanam, ``A time series is worth 64 words: Long-term forecasting with transformers,'' in \emph{ICLR}, 2023.

\bibitem{2017GCN}
T.~N. Kipf and M.~Welling, ``Semi-supervised classification with graph convolutional networks,'' in \emph{ICLR}, 2017.

\bibitem{2014GRU}
K.~Cho, B.~van Merrienboer, D.~Bahdanau, and Y.~Bengio, ``On the properties of neural machine translation: Encoder-decoder approaches,'' in \emph{SSST@EMNLP}.\hskip 1em plus 0.5em minus 0.4em\relax Association for Computational Linguistics, 2014.

\bibitem{2016TCN}
F.~Yu and V.~Koltun, ``Multi-scale context aggregation by dilated convolutions,'' in \emph{ICLR}, 2016.

\bibitem{2020GMAN}
C.~Zheng, X.~Fan, C.~Wang, and J.~Qi, ``{GMAN:} {A} graph multi-attention network for traffic prediction,'' in \emph{AAAI}, 2020.

\bibitem{HUTFormer}
Z.~Shao, F.~Wang, Z.~Zhang, Y.~Fang, G.~Jin, and Y.~Xu, ``Hutformer: Hierarchical u-net transformer for long-term traffic forecasting,'' \emph{arXiv preprint arXiv:2307.14596}, 2023.

\bibitem{2020STSGCN}
C.~Song, Y.~Lin, S.~Guo, and H.~Wan, ``Spatial-temporal synchronous graph convolutional networks: A new framework for spatial-temporal network data forecasting,'' in \emph{Proceedings of the AAAI}, 2020.

\bibitem{2020StemGNN}
D.~Cao, Y.~Wang, J.~Duan, C.~Zhang, X.~Zhu, C.~Huang, Y.~Tong, B.~Xu, J.~Bai, J.~Tong, and Q.~Zhang, ``Spectral temporal graph neural network for multivariate time-series forecasting,'' in \emph{NeurIPS}, 2020.

\bibitem{2021Z-GCNets}
Y.~Chen, I.~Segovia{-}Dominguez, and Y.~R. Gel, ``Z-gcnets: Time zigzags at graph convolutional networks for time series forecasting,'' in \emph{ICML}, vol. 139.\hskip 1em plus 0.5em minus 0.4em\relax {PMLR}, 2021.

\bibitem{STGODE}
Z.~Fang, Q.~Long, G.~Song, and K.~Xie, ``Spatial-temporal graph {ODE} networks for traffic flow forecasting,'' in \emph{SIGKDD}.\hskip 1em plus 0.5em minus 0.4em\relax {ACM}, 2021.

\bibitem{2023DLinear}
A.~Zeng, M.~Chen, L.~Zhang, and Q.~Xu, ``Are transformers effective for time series forecasting?'' in \emph{AAAI}, 2023.

\bibitem{DSformer}
C.~Yu, F.~Wang, Z.~Shao, T.~Sun, L.~Wu, and Y.~Xu, ``Dsformer: A double sampling transformer for multivariate time series long-term prediction,'' in \emph{CIKM}, 2023.

\bibitem{2022TimesNet}
H.~Wu, T.~Hu, Y.~Liu, H.~Zhou, J.~Wang, and M.~Long, ``Timesnet: Temporal 2d-variation modeling for general time series analysis,'' in \emph{ICLR}, 2023.

\bibitem{2022SimST}
X.~Liu, Y.~Liang, C.~Huang, H.~Hu, Y.~Cao, B.~Hooi, and R.~Zimmermann, ``Do we really need graph neural networks for traffic forecasting?'' \emph{CoRR}, vol. abs/2301.12603, 2023.

\bibitem{2022STID}
Z.~Shao, Z.~Zhang, F.~Wang, W.~Wei, and Y.~Xu, ``Spatial-temporal identity: {A} simple yet effective baseline for multivariate time series forecasting,'' in \emph{CIKM}.\hskip 1em plus 0.5em minus 0.4em\relax {ACM}, 2022.

\bibitem{2021STNorm}
J.~Deng, X.~Chen, R.~Jiang, X.~Song, and I.~W. Tsang, ``St-norm: Spatial and temporal normalization for multi-variate time series forecasting,'' in \emph{SIGKDD}.\hskip 1em plus 0.5em minus 0.4em\relax {ACM}, 2021.

\bibitem{LTSF_Survey}
Z.~Chen, M.~Ma, T.~Li, H.~Wang, and C.~Li, ``Long sequence time-series forecasting with deep learning: A survey,'' \emph{Information Fusion}, vol.~97, p. 101819, 2023.

\bibitem{ARIMA}
A.~A. Ariyo, A.~O. Adewumi, and C.~K. Ayo, ``Stock price prediction using the {ARIMA} model,'' in \emph{UKSim 2014}.\hskip 1em plus 0.5em minus 0.4em\relax {IEEE}, 2014.

\bibitem{ETS}
E.~S. Gardner~Jr, ``Exponential smoothing: The state of the art,'' \emph{Journal of forecasting}, vol.~4, no.~1, 1985.

\bibitem{GBRT}
J.~H. Friedman, ``Greedy function approximation: a gradient boosting machine,'' \emph{Annals of statistics}, 2001.

\bibitem{SVR}
H.~Drucker, C.~J.~C. Burges, L.~Kaufman, A.~J. Smola, and V.~Vapnik, ``Support vector regression machines,'' in \emph{NeurIPS}, 1996.

\bibitem{innovation}
Y.~Xu, X.~Liu, X.~Cao, C.~Huang, E.~Liu, S.~Qian, X.~Liu, Y.~Wu, F.~Dong, C.-W. Qiu \emph{et~al.}, ``Artificial intelligence: A powerful paradigm for scientific research,'' \emph{The Innovation}, vol.~2, no.~4, 2021.

\bibitem{innovation2}
F.~Wang, D.~Yao, Y.~Li, T.~Sun, and Z.~Zhang, ``Ai-enhanced spatial-temporal data-mining technology: New chance for next-generation urban computing,'' \emph{The Innovation}, vol.~4, no.~2, 2023.

\bibitem{2014Seq2Seq}
I.~Sutskever, O.~Vinyals, and Q.~V. Le, ``Sequence to sequence learning with neural networks,'' in \emph{NeurIPS}, 2014.

\bibitem{2015convLSTM}
X.~Shi, Z.~Chen, H.~Wang, D.~Yeung, W.~Wong, and W.~Woo, ``Convolutional {LSTM} network: {A} machine learning approach for precipitation nowcasting,'' in \emph{NeurIPS}, 2015.

\bibitem{2018DMVST}
H.~Yao, F.~Wu, J.~Ke, X.~Tang, Y.~Jia, S.~Lu, P.~Gong, J.~Ye, and Z.~Li, ``Deep multi-view spatial-temporal network for taxi demand prediction,'' in \emph{AAAI}, 2018.

\bibitem{2016GCN}
M.~Defferrard, X.~Bresson, and P.~Vandergheynst, ``Convolutional neural networks on graphs with fast localized spectral filtering,'' in \emph{NeurIPS}, 2016.

\bibitem{2019STMetaNet}
Z.~Pan, Y.~Liang, W.~Wang, Y.~Yu, Y.~Zheng, and J.~Zhang, ``Urban traffic prediction from spatio-temporal data using deep meta learning,'' in \emph{SIGKDD}.\hskip 1em plus 0.5em minus 0.4em\relax {ACM}, 2019.

\bibitem{2021DGCRN}
F.~Li, J.~Feng, H.~Yan, G.~Jin, F.~Yang, F.~Sun, D.~Jin, and Y.~Li, ``Dynamic graph convolutional recurrent network for traffic prediction: Benchmark and solution,'' \emph{ACM Transactions on Knowledge Discovery from Data}, vol.~17, no.~1, 2023.

\bibitem{jin2022automated}
G.~Jin, F.~Li, J.~Zhang, M.~Wang, and J.~Huang, ``Automated dilated spatio-temporal synchronous graph modeling for traffic prediction,'' \emph{IEEE Transactions on Intelligent Transportation Systems}, vol.~24, 2023.

\bibitem{2021ASTGNN}
S.~Guo, Y.~Lin, H.~Wan, X.~Li, and G.~Cong, ``Learning dynamics and heterogeneity of spatial-temporal graph data for traffic forecasting,'' \emph{TKDE}, vol.~34, 2022.

\bibitem{nas1}
X.~Wu, D.~Zhang, M.~Zhang, C.~Guo, B.~Yang, and C.~S. Jensen, ``Autocts+: Joint neural architecture and hyperparameter search for correlated time series forecasting,'' \emph{Proceedings of the ACM on Management of Data}, vol.~1, no.~1, 2023.

\bibitem{2020AGCRN}
L.~Bai, L.~Yao, C.~Li, X.~Wang, and C.~Wang, ``Adaptive graph convolutional recurrent network for traffic forecasting,'' in \emph{NeurIPS}, 2020.

\bibitem{2020MTGNN}
Z.~Wu, S.~Pan, G.~Long, J.~Jiang, X.~Chang, and C.~Zhang, ``Connecting the dots: Multivariate time series forecasting with graph neural networks,'' in \emph{SIGKDD}.\hskip 1em plus 0.5em minus 0.4em\relax {ACM}, 2020.

\bibitem{2021GTS}
C.~Shang, J.~Chen, and J.~Bi, ``Discrete graph structure learning for forecasting multiple time series,'' in \emph{ICLR}, 2021.

\bibitem{DFDGCN}
Y.~Li, Z.~Shao, Y.~Xu, Q.~Qiu, Z.~Cao, and F.~Wang, ``Dynamic frequency domain graph convolutional network for traffic forecasting,'' in \emph{ICASSP}.\hskip 1em plus 0.5em minus 0.4em\relax IEEE, 2024.

\bibitem{2022STEP}
Z.~Shao, Z.~Zhang, F.~Wang, and Y.~Xu, ``Pre-training enhanced spatial-temporal graph neural network for multivariate time series forecasting,'' in \emph{SIGKDD}.\hskip 1em plus 0.5em minus 0.4em\relax {ACM}, 2022.

\bibitem{2023NonGCN1}
J.~Deng, X.~Chen, R.~Jiang, D.~Yin, Y.~Yang, X.~Song, and I.~W. Tsang, ``Disentangling structured components: Towards adaptive, interpretable and scalable time series forecasting,'' \emph{IEEE Transactions on Knowledge and Data Engineering}, 2024.

\bibitem{2023NonGCN2}
H.~Liu, Z.~Dong, R.~Jiang, J.~Deng, J.~Deng, Q.~Chen, and X.~Song, ``Spatio-temporal adaptive embedding makes vanilla transformer sota for traffic forecasting,'' in \emph{CIKM}.\hskip 1em plus 0.5em minus 0.4em\relax {ACM}, 2023.

\bibitem{2023GraphFree}
X.~Wang, P.~Gu, P.~Wang, B.~Wang, Z.~Zhou, L.~Bai, and Y.~Wang, ``Graph-free learning in graph-structured data: {A} more efficient and accurate spatiotemporal learning perspective,'' \emph{CoRR}, vol. abs/2301.11742, 2023.

\bibitem{STNorm2}
J.~Deng, X.~Chen, R.~Jiang, X.~Song, and I.~W. Tsang, ``A multi-view multi-task learning framework for multi-variate time series forecasting,'' \emph{IEEE Transactions on Knowledge and Data Engineering}, vol.~35, no.~8, pp. 7665--7680, 2022.

\bibitem{2022MSDR}
D.~Liu, J.~Wang, S.~Shang, and P.~Han, ``{MSDR:} multi-step dependency relation networks for spatial temporal forecasting,'' in \emph{SIGKDD}.\hskip 1em plus 0.5em minus 0.4em\relax {ACM}, 2022.

\bibitem{2021LibCity}
J.~Wang, J.~Jiang, W.~Jiang, C.~Li, and W.~X. Zhao, ``Libcity: An open library for traffic prediction,'' in \emph{{SIGSPATIAL} '21: 29th International Conference on Advances in Geographic Information Systems, Virtual Event / Beijing, China, November 2-5, 2021}.\hskip 1em plus 0.5em minus 0.4em\relax {ACM}, 2021.

\bibitem{2021DL-Traff}
R.~Jiang, D.~Yin, Z.~Wang, Y.~Wang, J.~Deng, H.~Liu, Z.~Cai, J.~Deng, X.~Song, and R.~Shibasaki, ``Dl-traff: Survey and benchmark of deep learning models for urban traffic prediction,'' in \emph{CIKM}.\hskip 1em plus 0.5em minus 0.4em\relax {ACM}, 2021.

\bibitem{BasicTS}
Y.~Liang, Z.~Shao, F.~Wang, Z.~Zhang, T.~Sun, and Y.~Xu, ``Basicts: An open source fair multivariate time series prediction benchmark,'' in \emph{International Symposium on Benchmarking, Measuring and Optimization}.\hskip 1em plus 0.5em minus 0.4em\relax Springer, 2022.

\bibitem{jin2023spatio}
G.~Jin, Y.~Liang, Y.~Fang, J.~Huang, J.~Zhang, and Y.~Zheng, ``Spatio-temporal graph neural networks for predictive learning in urban computing: A survey,'' \emph{arXiv preprint arXiv:2303.14483}, 2023.

\bibitem{2021STFGCN}
M.~Li and Z.~Zhu, ``Spatial-temporal fusion graph neural networks for traffic flow forecasting,'' in \emph{AAAI}, vol.~35, no.~5, 2021.

\bibitem{2021SCINet}
M.~Liu, A.~Zeng, Z.~Xu, Q.~Lai, and Q.~Xu, ``Time series is a special sequence: Forecasting with sample convolution and interaction,'' \emph{arXiv preprint arXiv:2106.09305}, vol.~1, no.~9, 2021.

\bibitem{2022STG-NCDE}
J.~Choi, H.~Choi, J.~Hwang, and N.~Park, ``Graph neural controlled differential equations for traffic forecasting,'' in \emph{AAAI}, 2022.

\bibitem{NBeats}
B.~N. Oreshkin, D.~Carpov, N.~Chapados, and Y.~Bengio, ``N-beats: Neural basis expansion analysis for interpretable time series forecasting,'' in \emph{ICLR}, 2019.

\bibitem{Predictability1}
P.~Xu, L.~Yin, Z.~Yue, and T.~Zhou, ``On predictability of time series,'' \emph{Physica A: Statistical Mechanics and its Applications}, vol. 523, 2019.

\bibitem{Predictability2}
J.~Garland, R.~James, and E.~Bradley, ``Model-free quantification of time-series predictability,'' \emph{Physical Review E}, vol.~90, no.~5, 2014.

\bibitem{tsne}
L.~Van~der Maaten and G.~Hinton, ``Visualizing data using t-sne.'' \emph{Journal of machine learning research}, vol.~9, no.~11, 2008.

\bibitem{kde}
Y.-C. Chen, ``A tutorial on kernel density estimation and recent advances,'' \emph{Biostatistics \& Epidemiology}, vol.~1, no.~1, 2017.

\bibitem{homophily1}
J.~Zhu, R.~A. Rossi, A.~Rao, T.~Mai, N.~Lipka, N.~K. Ahmed, and D.~Koutra, ``Graph neural networks with heterophily,'' in \emph{AAAI}, 2021.

\bibitem{homophily2}
J.~Zhu, Y.~Yan, L.~Zhao, M.~Heimann, L.~Akoglu, and D.~Koutra, ``Beyond homophily in graph neural networks: Current limitations and effective designs,'' in \emph{NeurIPS}, 2020.

\bibitem{FPP}
R.~J. Hyndman and G.~Athanasopoulos, \emph{Forecasting: principles and practice}.\hskip 1em plus 0.5em minus 0.4em\relax OTexts, 2018.

\bibitem{2019ASTGCN}
S.~Guo, Y.~Lin, N.~Feng, C.~Song, and H.~Wan, ``Attention based spatial-temporal graph convolutional networks for traffic flow forecasting,'' in \emph{AAAI}, 2019.

\bibitem{2023LargeST}
X.~Liu, Y.~Xia, Y.~Liang, J.~Hu, Y.~Wang, L.~Bai, C.~Huang, Z.~Liu, B.~Hooi, and R.~Zimmermann, ``Largest: A benchmark dataset for large-scale traffic forecasting,'' \emph{NeurIPS}, 2024.

\bibitem{Triformer}
R.~Cirstea, C.~Guo, B.~Yang, T.~Kieu, X.~Dong, and S.~Pan, ``Triformer: Triangular, variable-specific attentions for long sequence multivariate time series forecasting,'' in \emph{IJCAI}, 2022.

\bibitem{LightGBM}
G.~Ke, Q.~Meng, T.~Finley, T.~Wang, W.~Chen, W.~Ma, Q.~Ye, and T.-Y. Liu, ``Lightgbm: A highly efficient gradient boosting decision tree,'' \emph{NeurIPS}, vol.~30, 2017.

\bibitem{DeepAR}
D.~Salinas, V.~Flunkert, J.~Gasthaus, and T.~Januschowski, ``Deepar: Probabilistic forecasting with autoregressive recurrent networks,'' \emph{International Journal of Forecasting}, vol.~36, no.~3, 2020.

\bibitem{NHiTS}
C.~Challu, K.~G. Olivares, B.~N. Oreshkin, F.~G. Ram{\'{\i}}rez, M.~M. Canseco, and A.~Dubrawski, ``{NHITS:} neural hierarchical interpolation for time series forecasting,'' in \emph{AAAI}, 2023.

\bibitem{M4Paper}
S.~Makridakis, E.~Spiliotis, and V.~Assimakopoulos, ``The m4 competition: 100,000 time series and 61 forecasting methods,'' \emph{International Journal of Forecasting}, vol.~36, no.~1, 2020.

\bibitem{2021ViT}
A.~Dosovitskiy, L.~Beyer, A.~Kolesnikov, D.~Weissenborn, X.~Zhai, T.~Unterthiner, M.~Dehghani, M.~Minderer, G.~Heigold, S.~Gelly, J.~Uszkoreit, and N.~Houlsby, ``An image is worth 16x16 words: Transformers for image recognition at scale,'' in \emph{ICLR}, 2021.

\bibitem{InnovationGeo}
T.~Zhao, S.~Wang, C.~Ouyang, M.~Chen, C.~Liu, J.~Zhang, L.~Yu, F.~Wang, Y.~Xie, J.~Li \emph{et~al.}, ``Artificial intelligence for geoscience: Progress, challenges and perspectives,'' \emph{The Innovation}, 2024.

\bibitem{AI4Science}
Y.~Xu, F.~Wang, Z.~An, Q.~Wang, and Z.~Zhang, ``Artificial intelligence for science—bridging data to wisdom,'' \emph{The Innovation}, vol.~4, no.~6, 2023.

\bibitem{MGSFformer}
C.~Yu, F.~Wang, Y.~Wang, Z.~Shao, T.~Sun, D.~Yao, and Y.~Xu, ``Mgsfformer: A multi-granularity spatiotemporal fusion transformer for air quality prediction,'' \emph{Information Fusion}, 2024.

\bibitem{Image1}
C.~Yang, H.~Zhou, Z.~An, X.~Jiang, Y.~Xu, and Q.~Zhang, ``Cross-image relational knowledge distillation for semantic segmentation,'' in \emph{CVPR}, 2022.

\bibitem{Image2}
L.~Huang, Y.~Zeng, C.~Yang, Z.~An, B.~Diao, and Y.~Xu, ``etag: Class-incremental learning via embedding distillation and task-oriented generation,'' in \emph{AAAI}, 2024.

\bibitem{GinAR}
C.~Yu, F.~Wang, Z.~Shao, T.~Qian, Z.~Zhang, W.~Wei, and Y.~Xu, ``Ginar: An end-to-end multivariate time series forecasting model suitable for variable missing,'' \emph{arXiv preprint arXiv:2405.11333}, 2024.

\bibitem{M4team2018dataset}
\BIBentryALTinterwordspacing
{Spyros Makridakis}, ``{M4} dataset,'' 2018. [Online]. Available: \url{https://github.com/M4Competition/M4-methods/tree/master/Dataset}
\BIBentrySTDinterwordspacing

\end{thebibliography}

\vspace{-1.2cm}

\begin{IEEEbiography}[{\includegraphics[width=1in,height=1.25in,clip,keepaspectratio]{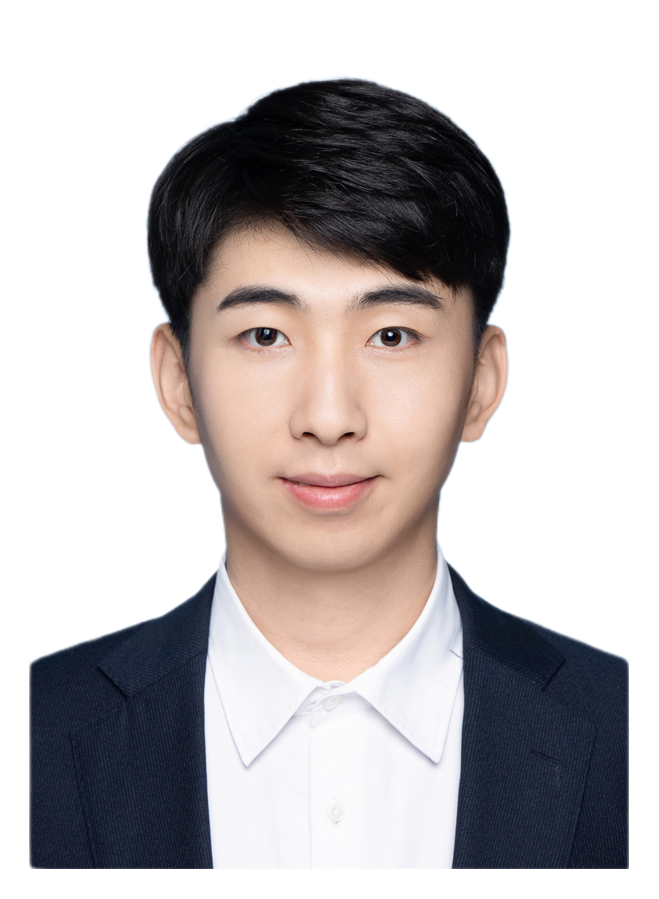}}]
  {Zezhi Shao} is an Assistant Research Fellow in Institute of Computing Technology, Chinese Academy of Sciences.
  He received the PhD degree in the Institute of Computing Technology,  Chinese Academy of Sciences in 2024.
  He received the B.E. degree from Shandong University, Jinan, China, in 2019.
  His research interests include multivariate time series forecasting, spatial-temporal data mining, and graph neural networks.
  He has published many papers as the first author in top journals and conferences such as TKDE, SIGKDD, VLDB, CIKM, etc.
\end{IEEEbiography}
\vspace{-1.2cm}
\begin{IEEEbiography}[{\includegraphics[width=1in,height=1.25in,clip,keepaspectratio]{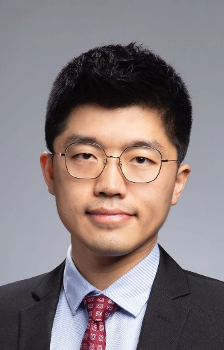}}]
  {Fei Wang}, born in 1988, PhD, associate professor. He received the PhD degree in computer architecture from Institute of Computing Technology, Chinese Academy of Sciences in 2017. From 2017 to 2020, he was a research assistant with the Institute of Technology, Chinese Academy of Sciences. Since 2020, he has been working as associate professor in Institute of Computing Technology, Chinese Academy of Sciences. His main research interest includes spatiotemporal data mining, Information fusion, graph neural networks.
\end{IEEEbiography}
\vspace{-1.2cm}
\begin{IEEEbiography}[{\includegraphics[width=1in,height=1.25in,clip,keepaspectratio]{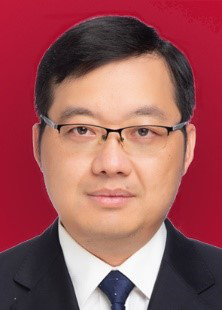}}]
  {Yongjun Xu} is a professor at Institute of Computing Technology, Chinese Academy of Sciences (ICT-CAS) in Beijing, China. He received his B.Eng. and Ph.D. degree in computer communication from Xi'an Institute of Posts \& Telecoms (China) in 2001 and Institute of Computing Technology, Chinese Academy of Sciences, Beijing, China in 2006, respectively. His current research interests include artificial intelligence systems, and big data processing. 
\end{IEEEbiography}
\vspace{-1.2cm}
\begin{IEEEbiography}[{\includegraphics[width=1in,height=1.25in,clip,keepaspectratio]{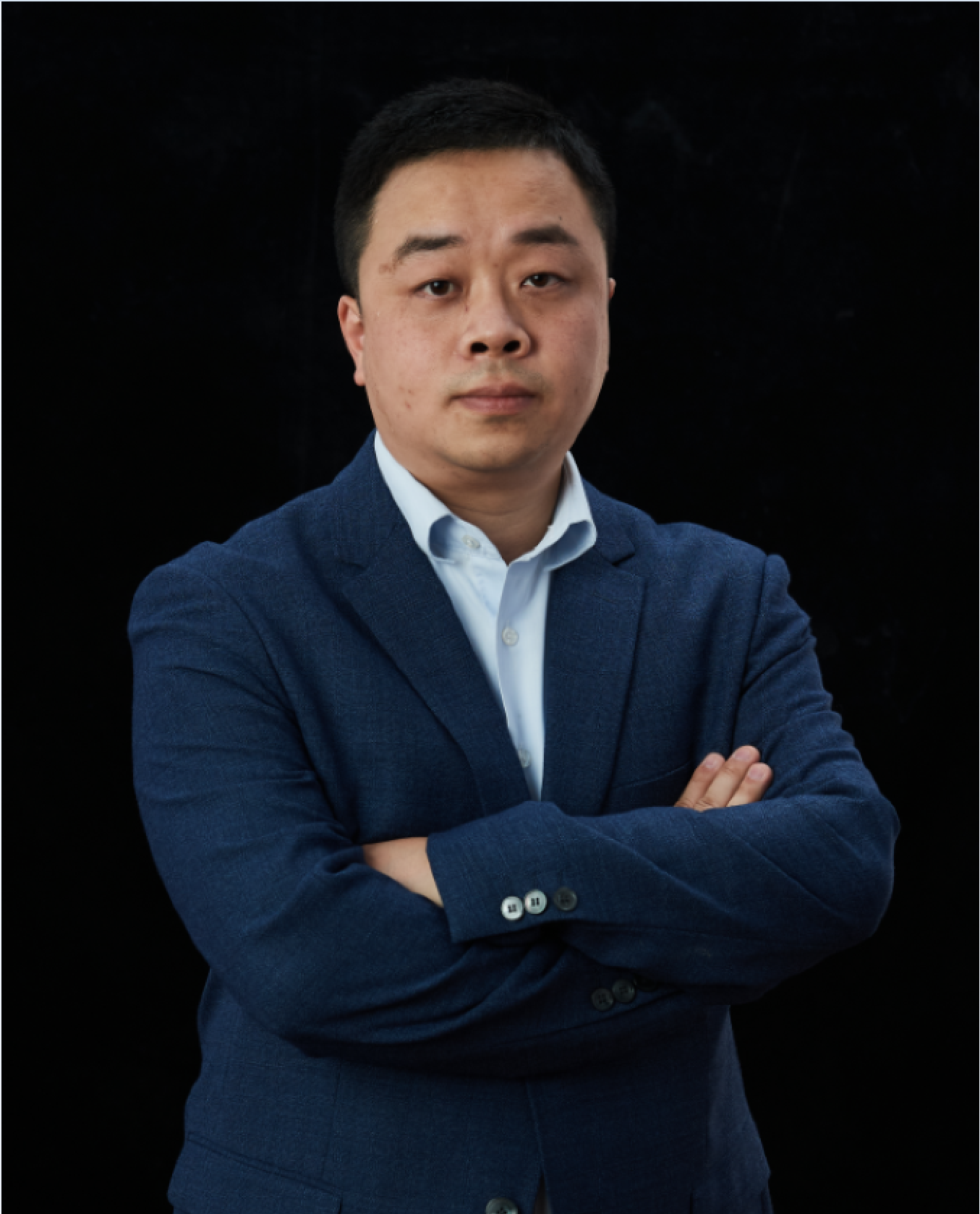}}]
  {Wei  Wei} received the PhD degree from the Huazhong University of Science and Technology, Wuhan, China, in 2012. He is currently a professor with the Department of Computer of Science and Technology, Huazhong University of Science and Technology. His current research interests include information retrieval, natural language processing, social computing and recommendation, cross-modal/multimodal computing,  deep learning, machine learning and artificial intelligence.
\end{IEEEbiography}
\vspace{-1.2cm}
\begin{IEEEbiography}[{\includegraphics[width=1in,height=1.25in,clip,keepaspectratio]{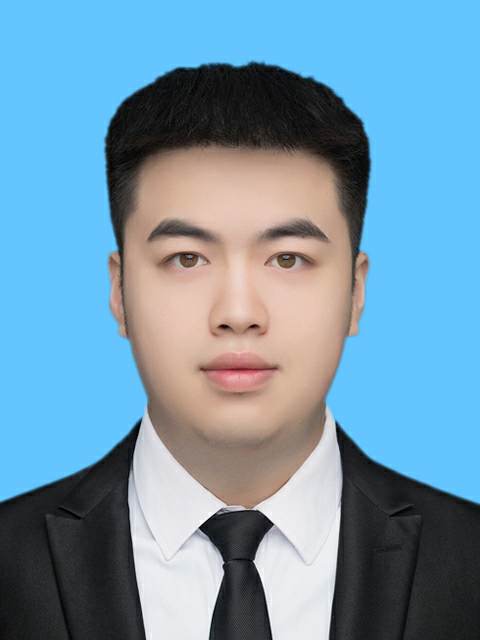}}]
  {Chengqing Yu} received the B.S. degree in Transport Equipment and Control Engineering from Central South University, Changsha, China, in 2019 and M.S. degree in Traffic and Transportation Engineering with Central South University, Changsha, China, in 2022, respectively. He is now PHD student in Institute of Computing Technology, Chinese Academy of Sciences, Beijing, China. His main research interests include deep learning, reinforcement learning, and time series forecasting.
\end{IEEEbiography}
\vspace{-1.2cm}

\begin{IEEEbiography}[{\includegraphics[width=1in,height=1.25in,clip,keepaspectratio]{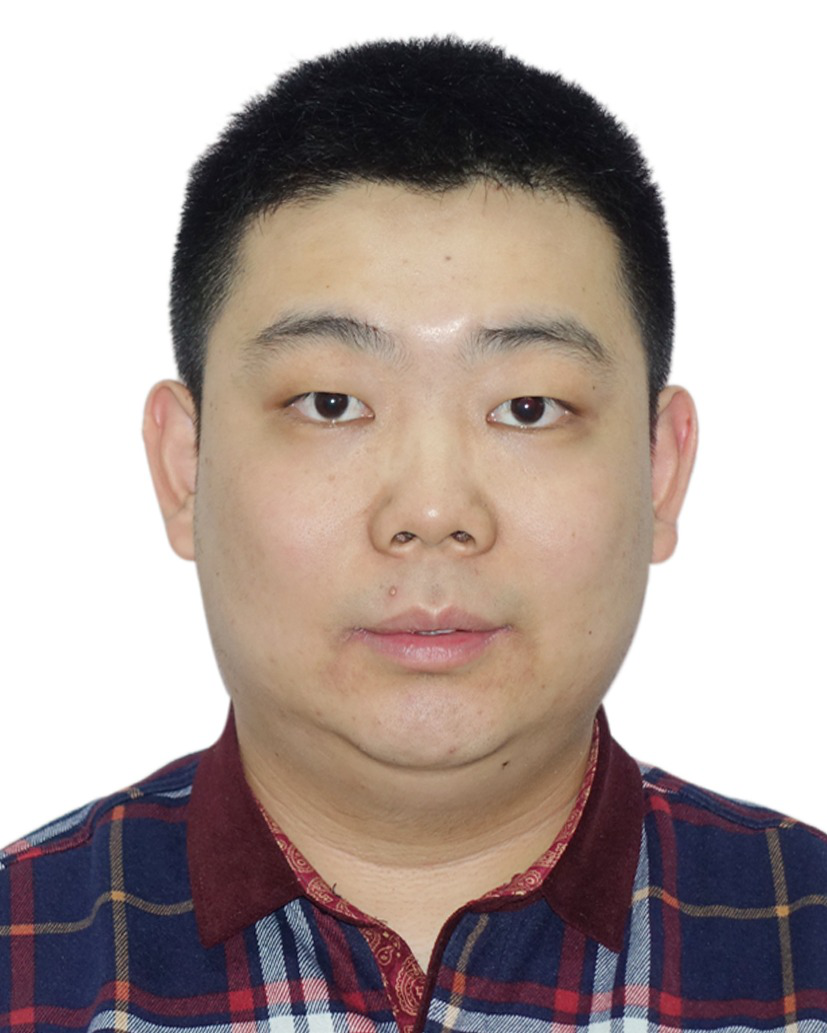}}]
  {Zhao Zhang} is currently an associate professor at the Institute of Computing Technology, Chinese Academy of Sciences, Beijing, China. He received the BE degree in computer science and technology from the Beijing Institute of Technology (BIT), Beijing, China, in 2015, and the PhD degree from the Institute of Computing Technology, Chinese Academy of Sciences, Beijing, China, in 2021. His research interests include data mining and applied machine learning, with a special focus on the representation and application of knowledge graphs.
\end{IEEEbiography}
\vspace{-1.2cm}

\begin{IEEEbiography}[{\includegraphics[width=1in,height=1.25in,clip,keepaspectratio]{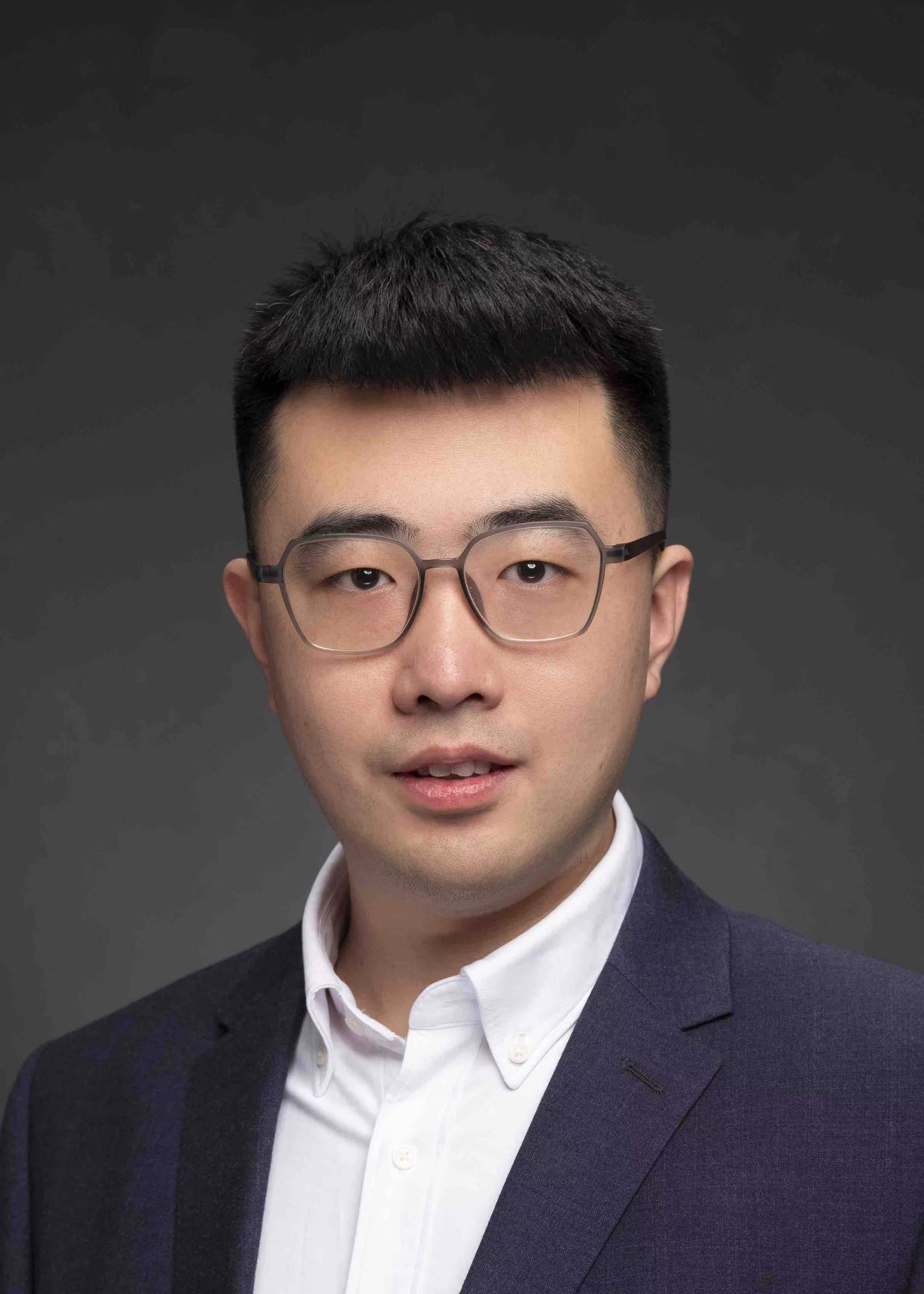}}]
  {Di Yao} is an associate professor in Institute of Computing Technology, Chinese Academy of Sciences (ICT, CAS). He obtained the Ph.D. degree in University of Chinese Academy of Sciences and conducted one year visiting at DMAL, Nanyang Technological University. His research interest lies in spatial temporal data mining, time series analysis, anomaly detection and causal discovery.
\end{IEEEbiography}
\vspace{-1.2cm}

\begin{IEEEbiography}[{\includegraphics[width=1in,height=1.25in,clip,keepaspectratio]{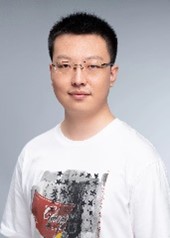}}]
  {Tao Sun} is an Assistant Research Fellow in Institute of Computing Technology, Chinese Academy of Sciences. He received the B.S. degree in information and computing science from the Xidian University, Xi’an, China, in 2016. He received the PhD degree in computer architecture from Institute of Computing Technology, Chinese Academy of Sciences in 2022. His research interest falls in the area of spatial-temporal data mining and trajectory data analysis. 
  He has published several papers in journals and conferences, such as CIKM, DASFAA, ICPR, etc.
\end{IEEEbiography}
\vspace{-1cm}

\begin{IEEEbiography}[{\includegraphics[width=1.0in,height=1.5in,clip,keepaspectratio]{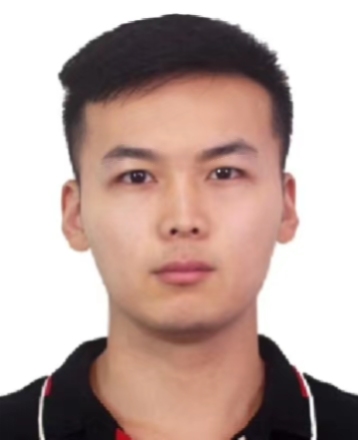}}]{Guangyin Jin} received the PhD degree from the College of Systems Engineering of National University of Defense Technology. His research interest falls in the area of spatial-temporal data mining, graph neural networks and urban computing. So far, he has published more than ten papers in JCR Q1-level international journals such as TKDE, TITS, TIST, TRC, INS, and top international conferences such as AAAI, CIKM, SIGSPATIL. 
\end{IEEEbiography}
\vspace{-1.2cm}

\begin{IEEEbiography}[{\includegraphics[width=1.0in,height=1.5in,clip,keepaspectratio]{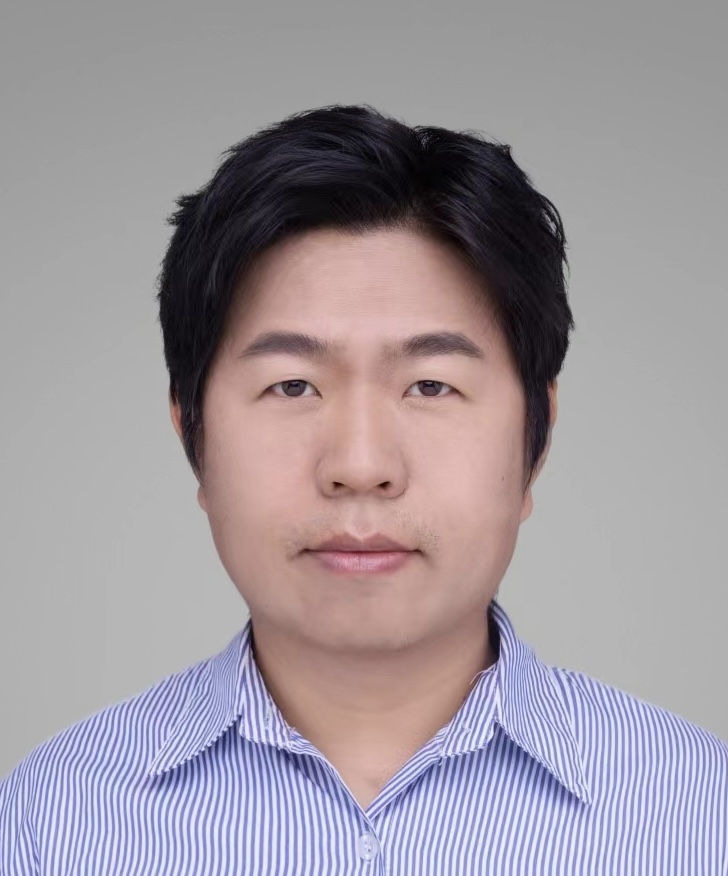}}]{Xin Cao} received the PhD degree from the School of Computer Science and Engineering, Nanyang Technological University, Singapore, in 2014. He is currently a senior lecturer with the School of Computer Science and Engineering, University of New South Wales, Sydney, Australia. His research interests include databases, data mining, and artificial intelligence.
\end{IEEEbiography}
\vspace{-1.2cm}

\begin{IEEEbiography}[{\includegraphics[width=1.0in,height=1.5in,clip,keepaspectratio]{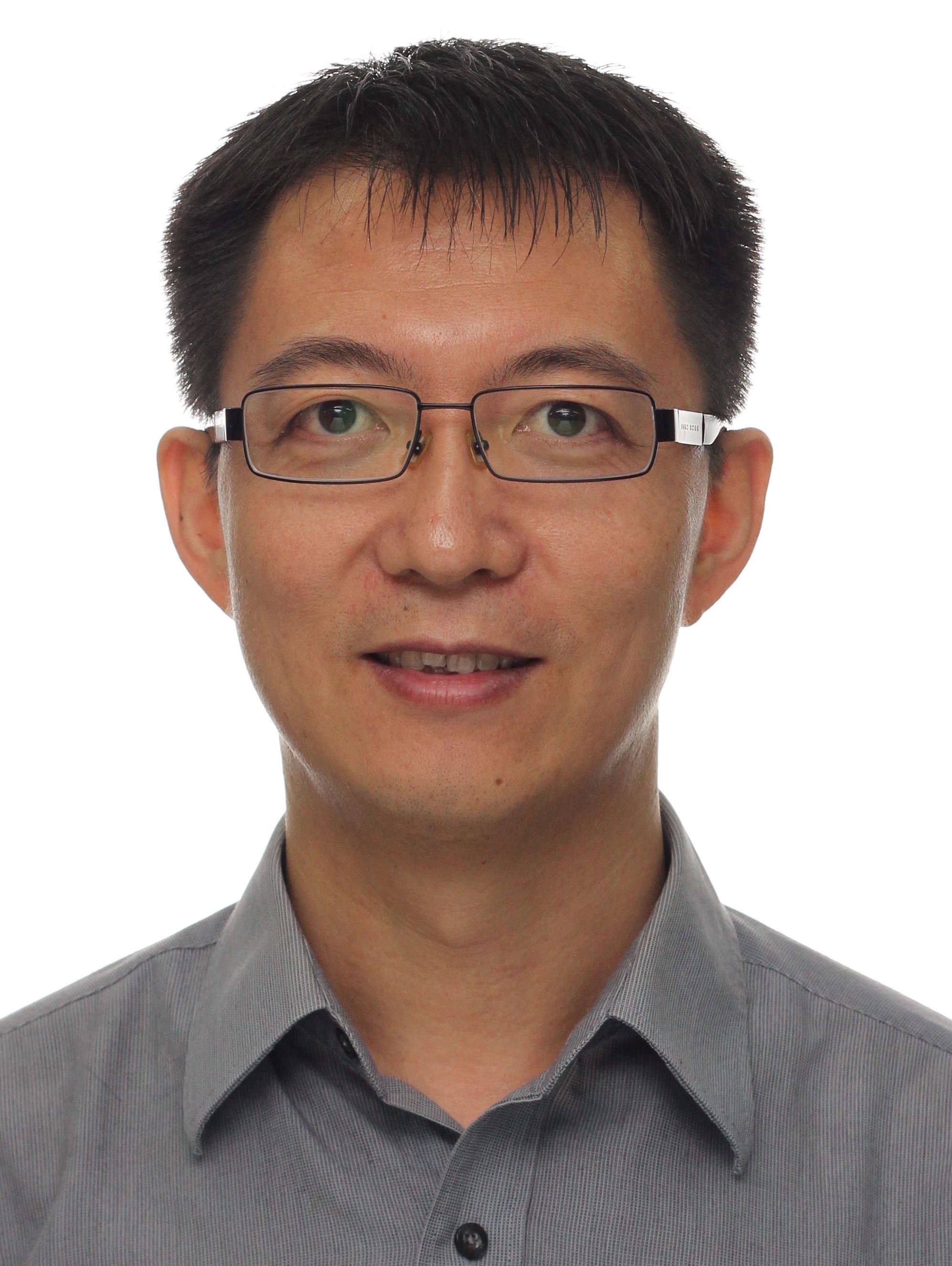}}]{Gao Cong} is currently a Professor in the School of Computer Science and Engineering at Nanyang Technological University (NTU). He previously worked at Aalborg University, Denmark, Microsoft Research Asia, and the University of Edinburgh. He received his PhD degree from the National University of Singapore in 2004. His current research interests include spatial data management, ML4DB, spatial-temporal data mining, and recommendation systems.
\end{IEEEbiography}
\vspace{-1cm}

\begin{IEEEbiography}
[{\includegraphics[width=1in,height=1.25in,clip,keepaspectratio]{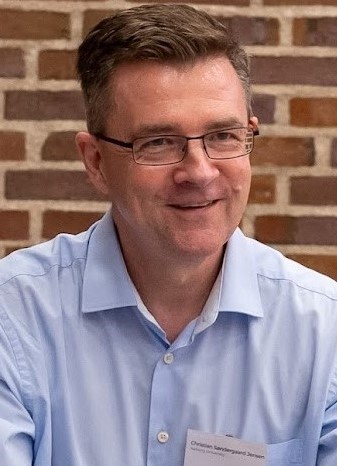}}]{Christian S. Jensen}
received the Ph.D. degree from Aalborg University, Denmark, where he is a professor. His research concerns data analytics and management with a focus on temporal and spatiotemporal data. He is a fellow of the ACM and IEEE, and he is a member of the Academia Europaea, the Royal Danish Academy of Sciences and Letters, and the Danish Academy of Technical Sciences.
\end{IEEEbiography}
\vspace{-1cm}

\begin{IEEEbiography}
[{\includegraphics[width=1in,height=1.25in,clip,keepaspectratio]{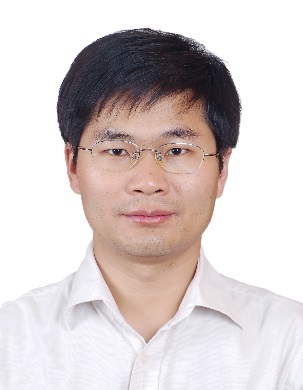}}]{Xueqi Cheng}
is a professor in the Institute of Computing Technology, Chinese Academy of Sciences (ICT-CAS) and the University of Chinese Academy of Sciences, and the director of the CAS Key Laboratory of Network Data Science and Technology. His main research interests include network science, web search and data mining, big data processing and distributed computing architecture. 
He has won the Best Paper Award in CIKM (2011), the Best Student Paper Award in SIGIR (2012), and the Best Paper Award Runner up of CIKM (2017).
\end{IEEEbiography}

\vfill

\end{document}